# Evaluating Large Language Models on the Spanish Medical Intern Resident (MIR) Examination 2024/2025:

# A Comparative Analysis of Clinical Reasoning and Knowledge Application

**Carlos Luengo Vera**
0009-0008-2591-1210
Universidad de Alcalá, Alcalá de Henares

**Ignacio Ferro Picón**
0009-0009-8593-9407
Binpar, Madrid

**M. Teresa del Val Nuñez**
0000-0001-6008-7935
Universidad de Alcalá, Alcalá de Henares

**José Andrés Gómez Gandía**
0009-0008-2591-1210
Universidad de Alcalá, Alcalá de Henares

**Antonio de Lucas Ancillo**
0000-0002-8876-7753
Universidad de Alcalá, Alcalá de Henares

**Víctor Ramos Arroyo**
0009-0004-3212-4179

**Carlos Milán Figueredo**
0009-0002-5024-3767

**Abstract**

**Purpose**

This study provides a comparative examination of large language models (LLMs) on the Spanish Medical Intern Resident (MIR) examination, focusing on both the 2024 and 2025 iterations. The MIR serves as a critical selection mechanism for medical graduates entering specialized training in Spain. A study is to be conducted on the ability of generative AI models to meet the challenges presented by MIR, with emphasis on clinical reasoning, image interpretation and epidemiological calculations. This research evaluates LLM performance in complex clinical scenarios and explores the extent to which LLMs demonstrate medical reasoning beyond mere information recall.

**Findings**

The results reveal key insights into the performance of 22 LLMs on the MIR 2024 and 2025 exams. The exam features 210 multiple-choice questions covering diverse medical domains and incorporates case-based scenarios, image interpretation (25 questions), and laboratory data analysis. A fine-tuned model, Miri Pro, achieved high accuracy in both years, scoring 195/210 in 2025. The analysis indicates a significant performance variation across models. Some LLMs exhibit greater proficiency in clinical reasoning, while others excel in knowledge recall, as evidenced by comparative scores on different question types. It is noted that general-based models without specialized medical adjustment tend to show lower accuracy, especially for questions requiring visual interpretation or Spain-specific epidemiological knowledge. The performance data shows differences in performance between 2024 and 2025, suggesting potential changes in exam difficulty or model training.

**Discussion**

The varying degrees of success among different LLMs on the MIR exam raise important questions about medical education and assessment. The ability of some models to correctly answer novel questions suggests reasoning capabilities beyond rote memorization. The incorporation of AI-based scoring in the MIR exam underscores the increasing role of AI in medical education. The findings suggest a need to re-evaluate traditional assessment methods and consider a greater emphasis on skills that AI currently struggles with, such as ethical judgment and nuanced clinical insight. AI could serve as a powerful tool for personalized learning, identifying knowledge gaps and providing tailored support for medical trainees, but requires careful ethical oversight. Also note the higher scores of actual test takers 186/210.

**Originality**

This study offers a novel contribution by providing a comparative assessment of a diverse set of generative AI models on both the 2024 and 2025 MIR examinations, including zero-shot evaluations of foundation models and fine-tuned systems. It highlights the importance of model architecture and fine-tuning for achieving high levels of accuracy on complex medical reasoning tasks. The inclusion of multimodal item analysis (image-based questions) further enhances the originality of the work. The study is contextualized by the Spanish healthcare system and the unique challenges of the MIR exam, providing insights relevant to policymakers, educators, and researchers seeking to integrate AI into medical training and assessment.

**Practical Implications**

The findings suggest a need for medical education to adapt to technological advances, potentially integrating AI literacy into curricula and emphasizing skills that are difficult to automate, such as clinical empathy and moral reasoning. Licensing bodies may consider updating exam formats to better assess the competencies required in modern medical practice. The effectiveness of fine-tuned models points to opportunities for creating AI tools that support both learning and clinical decision-making. Ethical frameworks and regulatory guidelines are needed to address issues of bias, data protection, and the responsible use of AI in medical education and practice. Continuous professional development is essential to ensure that medical practitioners can effectively collaborate with and critically evaluate AI tools.

**Keywords**

Medical Internship and Residency (MIR), Artificial Intelligence, Generative AI, Clinical Reasoning Assessment, Medical Licensing Examination, Spanish Healthcare System, Medical Education, AI in Healthcare.

**Introduction**

The Spanish Medical Internship and Residency (MIR) examination represents a high-stakes assessment, serving as the primary gateway for medical graduates seeking specialized training within the Spanish healthcare system. As generative artificial intelligence (AI) rapidly advances across various sectors, including healthcare, it becomes crucial to rigorously evaluate its capabilities within high-stakes medical assessments (Reddy, 2024). This study provides a comprehensive examination of large language models (LLMs) on the MIR, focusing on the 2024 and 2025 iterations. The MIR's design, which emphasizes complex clinical reasoning, multimodal elements (image interpretation), and Spain-specific epidemiological considerations, provides a robust testbed for evaluating the potential and limitations of these models.

This research is guided by two primary hypotheses. First, it is postulated that the performance level of LLMs in the MIR is closely related to their complex clinical reasoning. The MIR is not merely a test of factual recall; it demands that candidates apply medical knowledge to solve intricate, case-based clinical scenarios. Consequently, expert models in multistep problem solving and diagnostic reasoning are expected to outperform those focused primarily on information retrieval. Second, it aims to contrast the reasoning capabilities of different LLMs(Yuan et al., 2024) and their sensitivity to the quality and content of training data. A key question is whether LLMs demonstrate substantive medical reasoning, or if their apparent competence stems primarily from sophisticated memorization and pattern recognition within existing datasets (g Lema, 2023). This exploration is particularly timely given concerns about potential data contamination and the indexed information already available to these models. The 2025 MIR included two questions where answer options were modified to mitigate this concern.

The study's novelty lies in its comparative assessment of a diverse set of generative AI models on both the 2024 and 2025 MIR examinations, including zero-shot evaluations of foundation models and fine-tuned systems (Yu et al., 2023). By focusing on the Spanish healthcare system and the unique challenges of the MIR exam, this study provides contextual insights relevant to policymakers, educators, and researchers seeking to integrate AI into medical training and assessment.

Ultimately, this research aims to inform the future of medical education and licensing by providing empirical evidence on the capabilities and limitations of AI in the context of a rigorous, real-world medical examination. The findings contribute to a broader understanding of the transformative potential of AI in medical practice and training, while also underscoring the importance of ethical oversight and responsible implementation (Esmaeilzadeh, 2024).

Based on the analysis, two primary hypotheses were directly addressed:

1.The first hypothesis posited that LLM performance on the MIR examination is intrinsically linked to their capacity for complex clinical reasoning. In other words, models that excel at processing multifaceted, case-based clinical scenarios are expected to achieve higher accuracy on the exam.

2.The second hypothesis explored whether the performance differentials among LLMs stem primarily from genuine reasoning capabilities rather than simple memorization of information.

**2.Theoretical Framework**

The theoretical framework for understanding AI's role in medical education and practice is in a state of rapid evolution, reflecting the dynamic nature of both artificial intelligence and healthcare (Charow et al., 2021). The impressive performance of AI models on standardised exams like the MIR, coupled with advancements in AI reasoning capabilities, presents a landscape rich with both exciting opportunities and significant challenges. These advances are not mere academic curiosities, but harbingers of potential paradigm shift in the approach to medical education, clinical decision making and healthcare delivery.

As it moves through this period of transformation, it is increasingly crucial to foster interdisciplinary collaboration between AI researchers, medical professionals, ethicists, and policy makers. This collaborative approach will be essential in shaping the future integration of AI in medicine, ensuring that technological advancements are harnessed to enhance patient care, improve medical education, and support healthcare professionals in their critical work (Gandía et al., 2025). The ongoing refinement of these technologies must be guided by a commitment to ethical implementation, with a clear focus on augmenting human medical expertise rather than seeking to replace it. In this evolving landscape, the true potential of AI in medicine lies not in its ability to outperform humans on standardised tests, but in its capacity to work synergistically with human intelligence, compassion, and experience to elevate the standard of healthcare for all.

**What´s known**

Large Language Models (LLM) represent a milestone in the evolution of artificial intelligence (AI), enabling the generation of text with an unprecedented level of coherence and fluency. Their development has been made possible by advances in natural language processing (NLP), deep learning and access to large volumes of textual data. Since their inception, these models have demonstrated amazing capabilities in tasks such as machine translation, question answering and creative content generation, thus driving the generative artificial intelligence revolution.

The concept of language models is not new; however, their evolution into large-scale models has been driven by the emergence of more sophisticated architectures. From the first attempts with statistical models and recurrent neural networks (RNNs), to the emergence of Transformers in 2017 with the paper "Attention is All You Need" by(Vaswani et al., 2017)The PLN has undergone a radical transformation. Models such as GPT-3, BERT and T5 ushered in a new era where Transformer-based architectures demonstrated an unprecedented ability to understand and generate text.

Today, large language models (LLMs) have evolved into indispensable tools across a wide range of sectors. With the recent market introduction of Deepseek and Grok, alongside established models like GPT-4, Claude, LLaMA, and Gemini, these systems now demonstrate exceptional capabilities in semantic understanding, code generation, research assistance, and creative process automation. Their impact is evident in fields such as education, healthcare, law, and data science, where they streamline tasks like report writing, summary generation, and the analysis of large volumes of information.

Despite their potential, LLMs also face significant challenges, such as algorithmic bias, data hallucination (Okonji et al., 2024), and misuse of content generation. Regulation and ethics in the development of these models have become key issues to ensure responsible and equitable use of generative AI.

**What´s new**

The landscape of artificial intelligence in medical education has undergone a significant transformation in recent years, with generative AI models demonstrating remarkable capabilities in tackling complex medical examinations. The Spanish Medical Internship and Residency (MIR) exam, a rigorous assessment for medical graduates seeking specialisation in Spain, has become a notable benchmark for evaluating the reasoning and knowledge application abilities of AI models in a medical context.

In 2024, the performance of AI models on the MIR exam reached unprecedented levels, with GPT-4 achieving an impressive 82.4% accuracy rate (173 correct answers out of 210 questions). This marked a substantial improvement over its predecessor, ChatGPT-3, which scored 51.4% (108 correct answers) in the previous year's exam. The advancement was particularly notable in specialties such as Rheumatology, Paediatrics, Geriatrics, and Oncology, although some fields like Pulmonology and Ophthalmology showed less progress.

The success of GPT-4 in the 2024 MIR exam underscores the prompt evolution of AI capabilities in processing and analysing complex medical information. When the results were translated into the net score (accounting for incorrect answers), GPT-4 achieved a position between 1,100 and 1,300 among human candidates, placing it in the 90th to 92nd percentile. This performance level suggests that AI models are approaching, and in some cases surpassing, the capabilities of many human medical students in standardised testing scenarios.

However, it is crucial to note that while these results are impressive, they do not necessarily indicate that AI models possess the same level of understanding or clinical competence as human medical professionals. The MIR exam, while comprehensive, primarily tests factual knowledge and problem-solving skills in a controlled, multiple-choice format. It does not fully capture the nuanced decision-making, empathy, and hands-on skills required in real-world medical practice.

The quick advancement of AI models in medical examinations has been paralleled by developments in other areas of AI reasoning. OpenAI's introduction of the o3 and o3 mini models in late 2024 marked a significant milestone in AI's problem-solving capabilities, particularly in science, coding, and mathematics. These models, built upon the foundation of their o1 predecessors, promise enhanced reasoning abilities that go beyond simple pattern recognition or information retrieval.

The o3 models introduce a novel approach to AI reasoning, incorporating a "test-time compute" functionality that allows the system to consider multiple possible answers before committing to a response. This method of processing mimics human deliberation and has resulted in significantly improved performance on complex reasoning tasks. For instance, the o3 model achieved an impressive 87.5% score on the ARC-AGI test in its high-compute mode, far surpassing the 32% scored by its predecessor, o1.

These advancements in AI reasoning capabilities have profound implications for the field of medical education and practice. The ability of AI models to process vast amounts of medical literature, clinical guidelines, and patient data could potentially augment the decision-making processes of healthcare professionals. However, the integration of such advanced AI systems into medical practice raises important ethical and practical considerations.

One of the key challenges in the development and deployment of these advanced AI models is the balance between performance and accessibility. The high-compute version of o3 (Arrieta et al., 2025), while demonstrating superior reasoning capabilities, comes with substantial computational costs, with estimates suggesting a price of over $1,000 per task. This raises questions about the economic viability and equitable access to such powerful AI tools in medical education and practice.

The development of more accessible versions, such as the o3 mini, scheduled for release in early 2025, aims to address some of these concerns. The o3 mini is designed to offer enhanced reasoning capabilities while maintaining a more manageable computational footprint, potentially making it more suitable for widespread adoption in educational and clinical settings. The development of more affordable versions, such as the o3 mini, already in production since early 2025, is intended to address some of these issues.)

Looking ahead to the future of AI in medical education and practice, the field is at a critical juncture. The impressive performance of models like GPT-4(Aronson et al., 2024) on standardised medical exams and the promising capabilities of reasoning-focused models like o3 suggest that AI could play an increasingly significant role in supporting medical education, research, and clinical decision-making.

However, it is crucial to approach these developments with a balanced perspective. While AI models have demonstrated remarkable abilities in processing information and solving complex problems, they still lack the holistic understanding, ethical reasoning (Moulaei et al., 2024), and empathetic capabilities that are fundamental to medical practice (Abbasian et al., 2024). The challenge for the medical community will be to harness the power of these AI tools while maintaining the irreplaceable human elements of healthcare.

## 3. Methodology

This research employed a comparative cross-sectional design to investigate the performance of twenty-two Large Language Models (LLMs) on the Spanish Medical Intern Resident (MIR) examinations for 2024 and 2025. The MIR, widely regarded for its comprehensive coverage of clinical, diagnostic, and epidemiological competencies, served as a rigorous testbed for evaluating both general-purpose and domain-fine-tuned LLMs.

Initially, the official 2024 and 2025 MIR examinations were digitised in their original Spanish format. Each exam comprises 210 multiple-choice questions (MCQs), with around 25 items requiring the interpretation of images (e.g., radiological scans, ECGs, histopathology slides). Two questions in the 2025 exam were intentionally modified to reduce potential data leakage from sources already indexed by certain models. All MIR content was used strictly for research purposes and safeguarded to maintain confidentiality and exam integrity.

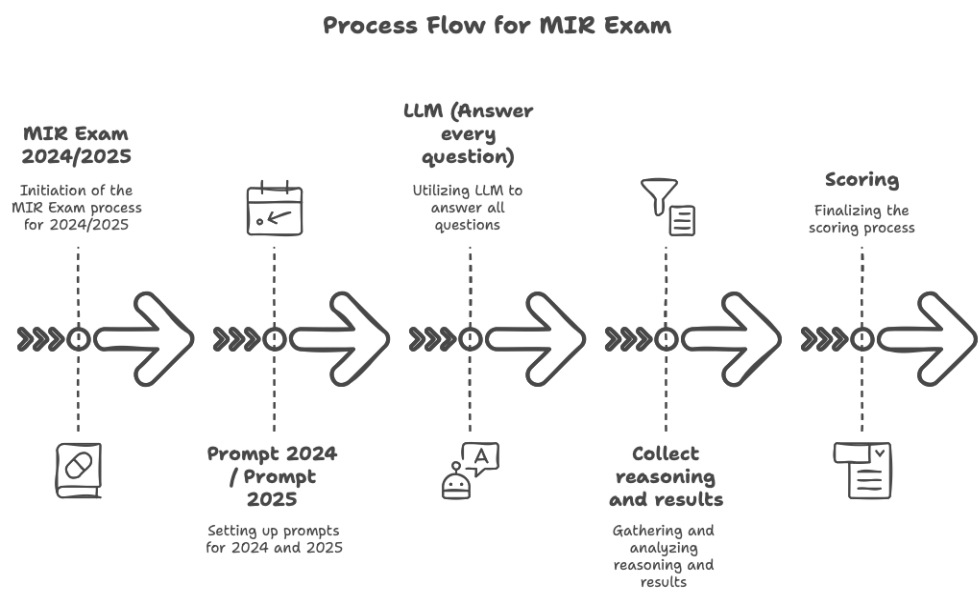

*Figure 1- Process Flow for MIR Exam*

A total of twenty-two LLMs were selected, ranging from high-profile general-purpose architectures (e.g., GPT-4 Turbo, Claude Sonet 3.5) to specialised models such as Miri Pro, which had undergone fine-tuning on Spanish medical corpora. To ensure consistency, every model was tested under zero-shot conditions, receiving no additional hints or clarifications beyond a standardised prompt. This prompt instructed the LLM to identify the single best answer (out of four choices) and provide a step-by-step explanation of its reasoning in Spanish. For multimodal-capable models, image-based questions were presented along with the relevant visual inputs. Text-only models are presented solely with the question, without any indication of an accompanying image, which prevents them from viewing or interpreting it. Instead, these models are supplied with written descriptions of the images that highlight the key visual features otherwise evident in the graphic.

Each LLM processed the 210 questions from both the 2024 and 2025 exams, generating responses that were automatically stored and timestamped. Correctness was determined by matching the final answer to the official solution key provided by the Spanish Ministry of Health, yielding a raw accuracy score (number of correct responses out of 210) that was later converted to a percentage. These results were supplemented by metadata on whether each question required image interpretation or tested domain-specific knowledge. A human performance reference—represented by the highest-achieving candidate in the 2025 MIR—served as a benchmark for contextualising the relative performance of each model.

To facilitate meaningful comparisons, accuracy data were subjected to basic descriptive statistics (mean, median, standard deviation). Further, models were grouped into categories (e.g., general-purpose vs. fine-tuned) for subgroup analyses. Paired tests were conducted to detect significant changes in individual models' performance from 2024 to 2025, while non-parametric methods were employed if the distributional assumptions for parametric tests were not met. Additionally, a series of omnibus tests was performed to ascertain whether performance differed significantly among multiple models, followed by post-hoc analyses where warranted.

Throughout the study, robust data management protocols were maintained, with model outputs stored in encrypted repositories. This controlled, consistent framework for measuring each model's capacity to address a demanding medical licensing exam lays the groundwork for further exploration into how LLMs might be integrated into future clinical or educational contexts.

## 4. Analysis and Results

### 4.1. Analysis of Large Language Model Performance on the MIR 2024 and 2025 Examinations

This section provides a detailed analysis of the performance of 22 Large Language Models (LLMs) evaluated on the Spanish Medical Internship and Residency (MIR) examinations for 2024 and 2025. The analysis focuses on two primary hypotheses: (1) the relationship between LLM performance and their ability to engage in complex clinical reasoning, and (2) the extent to which model performance reflects reasoning capabilities rather than memorisation. The results are contextualised within the evolving landscape of medical education and assessment, with a focus on the implications of AI integration into high-stakes medical licensing exams.

**Evaluation Overview**

The MIR examination is a high-stakes test designed to assess medical graduates' readiness for residency training in Spain. It includes 210 multiple-choice questions covering a wide range of medical disciplines, with an emphasis on clinical reasoning, case-based scenarios, image interpretation, and public health management. The evaluation involved zero-shot testing of LLMs using a standardised prompt in Spanish, ensuring consistency across all models.

**Prompt 2024 and 2025**

```
# Prompt MIR 2025
prompt_2025 = """
You are a health expert tasked with determining the correct answer to a question from the 2025 MIR exam.
The MIR exam is an entrance exam for medical residency in Spain, so all information will be focused on that country.
Each question will have four possible answers, but only one is correct.
Explain your reasoning step by step as you identify the correct answer in Spanish.
"""

# Prompt MIR 2024
prompt_2024 = """
You are a health expert tasked with determining the correct answer to a question from the 2024 MIR exam.
The MIR exam is an entrance exam for medical residency in Spain, so all information will be focused on that country.
Each question will have four possible answers, but only one is correct.
Explain your reasoning step by step as you identify the correct answer in Spanish.
"""
```

The dataset comprised both the 2024 and 2025 MIR exams. Each exam consists of 210 multiple-choice questions spanning a wide range of medical disciplines, including clinical reasoning, case-based scenarios, and public health management. Additionally, each exam features approximately 25 questions that require image interpretation, thereby assessing the models' multimodal capabilities.

**Overall Performance Comparison: 2024 vs. 2025**

The performance of LLMs varied significantly across models and years.

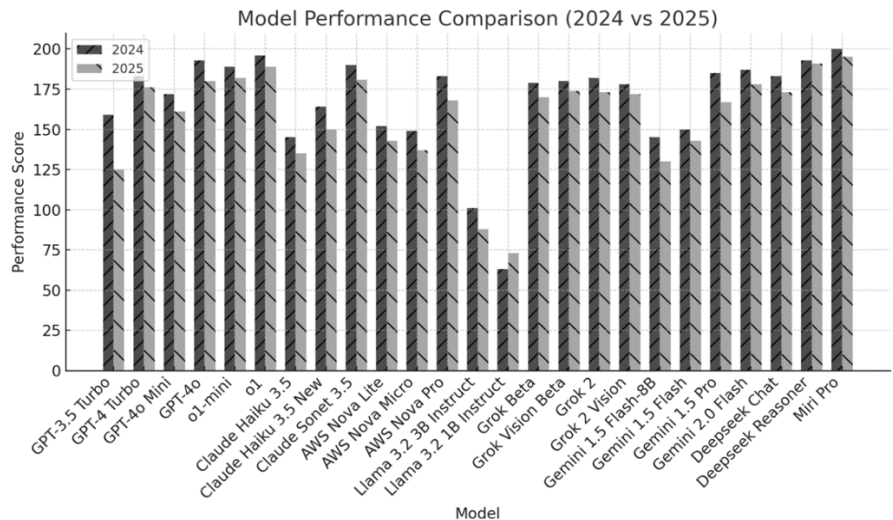

*Figure 2 Overall Performance Comparison: 2024 vs. 2025*

This graph should display the raw scores and normalised scores for each model in both years.

- **Top Performer**: Miri Pro achieved the highest scores in both years, with 200/210 (97.56%) in 2024 and 195/210 (95.59%) in 2025. Its consistent performance underscores its fine-tuning with domain-specific medical content.
- **Lowest Performers**: Llama models exhibited the weakest performance. For instance, Llama 3.2 3B Instruct scored 101/210 (49.27%) in 2024 and dropped to 88/210 (43.14%) in 2025.
- **Human Comparison**: The best human score in 2025 was 165/210 (78.57%), highlighting that several LLMs outperformed human candidates.

**Year-to-Year Comparison**

A comparison between the two years reveals interesting trends.

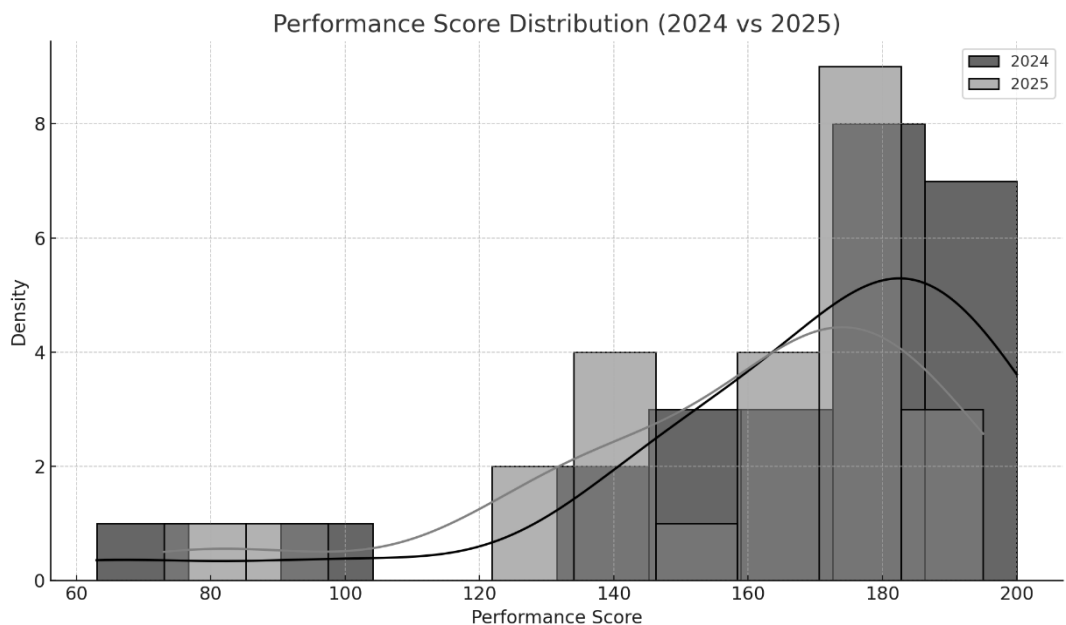

*Figure 3 Year-to-Year Comparison*

**Performance Change from 2024 to 2025**

This graph should illustrate changes in normalised scores between the two years for each model.

1. **Decline in Performance**: Most models experienced a slight decline from 2024 to 2025:

    - GPT-4 Turbo dropped from 89.27% to 86.27%.
    - Claude Sonet 3.5 decreased from 92.68% to 88.73%.
    - Miri Pro also saw a small reduction from 97.56% to 95.59%.

2. **Improvement in Some Models**: Interestingly, Llama 3.2 1B Instruct improved its score from 63/210 (30.73%) in 2024 to 73/210 (35.78%) in 2025, although its overall performance remained low.

3. **Impact of Modified Questions**: The slight decline across most models may reflect the introduction of modified questions in the 2025 exam, which were designed to test reasoning capabilities rather than reliance on memorised data.

**Reasoning vs Memorisation**

One of the central objectives was to evaluate whether LLMs demonstrated genuine reasoning or relied primarily on memorisation.

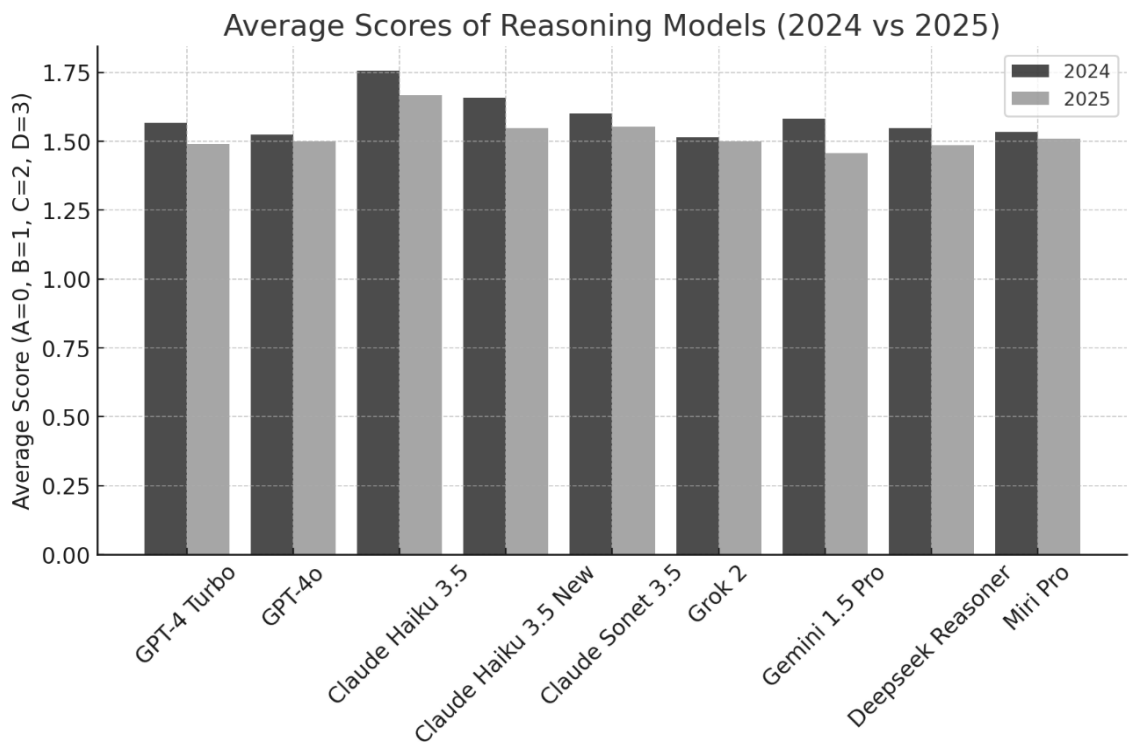

Figure 4 Reasoning vs Memorisation

- **Deepseek Reasoner**: This model excelled at reasoning tasks, achieving near-perfect scores (94.15% in 2024; 93.63% in 2025). Its architecture generates "reasoning tokens," which appear to enhance its ability to process complex clinical scenarios.
- **GPT-4o Mini vs GPT-4o**: While both models performed well overall, GPT-4o consistently outperformed GPT-4o Mini, suggesting that architectural differences contribute significantly to reasoning capabilities.
- **Modified Questions**: Models such as Miri Pro demonstrated strong generalisation abilities by correctly answering modified questions in the 2025 exam, further supporting their reasoning capabilities.

**Performance on Image-Based Questions**

The MIR includes a subset of questions requiring image interpretation, such as histological slides or diagnostic imaging.

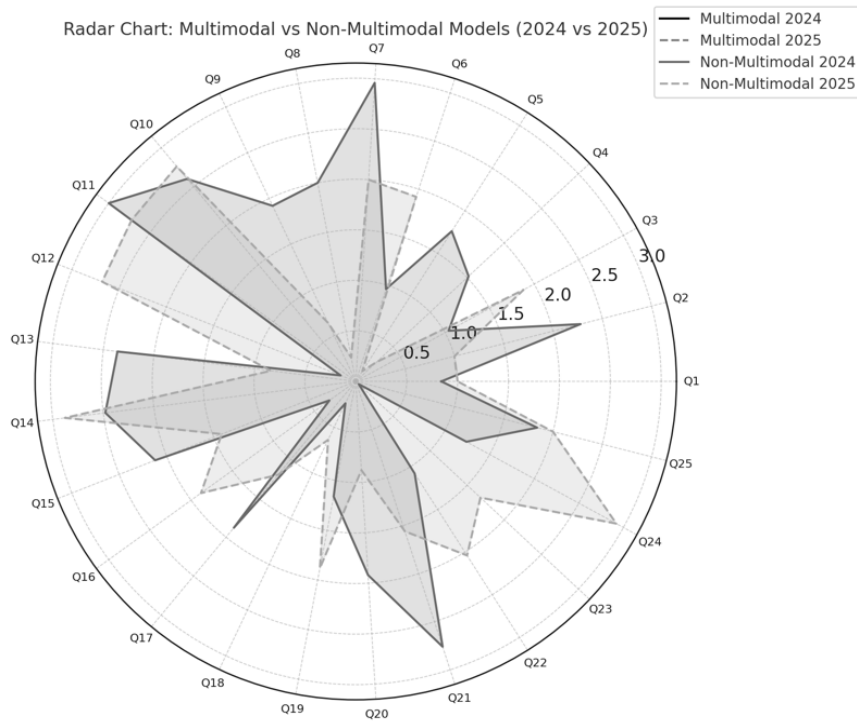

*Figure 5 Performance on Image-Based Questions*

This graph should display scores for image-based questions compared to other question types.

1. **Multimodal Models**: Models with visual processing capabilities, such as Grok Vision Beta and Gemini Vision Pro, performed better on these questions compared to text-only models.

- Grok Vision Beta scored consistently high across both years (87.80% in 2024; 85.29% in 2025).
- Text-only models like Claude Haiku struggled with these items, scoring below their average performance levels.

2. **Human Benchmark**: Human candidates demonstrated relatively strong performance on image-based questions but were still outperformed by multimodal AI models like Grok Vision Beta.

**Fine-Tuning and Domain-Specific Knowledge**

Fine-tuned models demonstrated a clear advantage over general-purpose LLMs.

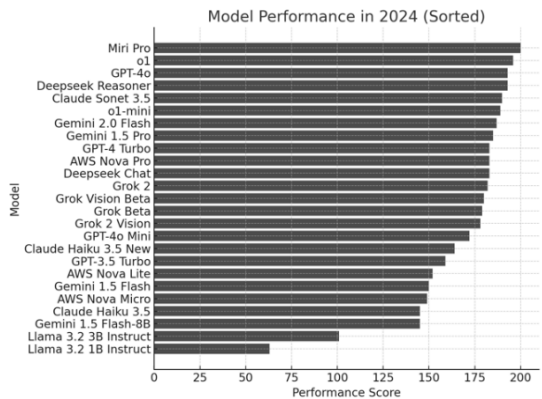
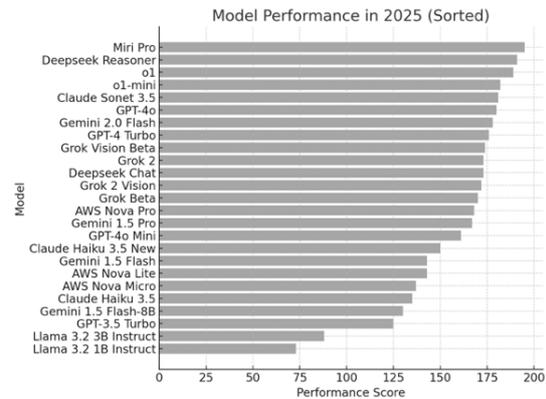

*Figure 6 Model Performance 2024*  *Figure 7 Model Performance 2025*

This graph should compare the normalised scores of fine-tuned models like Miri Pro against general-purpose models like GPT-3.5 Turbo.

- **Miri Pro's Success**: Fine-tuned with proprietary medical content, Miri Pro consistently outperformed all other models.
- **General-Purpose Models**: While general-purpose models like GPT-4 Turbo performed well overall (89.27% in 2024; 86.27% in 2025), they lagged fine-tuned counterparts on specialised tasks such as epidemiology questions specific to Spain.

**Human vs Machine**

The comparison between human candidates and AI models highlights key insights:

- **AI Superiority**: Several AI models outperformed human candidates, whose best score was 165/210 (78,57%) in the MIR 2025 exam.
- **Human Advantage**: Humans demonstrated strengths in ethical decision-making and nuanced clinical judgement—areas where AI still struggles.

**Key Observations**

1. **Complex Clinical Reasoning**:

- Models optimised for chain-of-thought reasoning consistently outperformed others.
- Fine-tuned systems like Miri Pro excelled at solving complex case-based scenarios.

2. **Impact of Training Data**:

- The inclusion of domain-specific content significantly enhanced model performance.

- General-purpose models showed limitations when confronted with highly specialised tasks.

3. **Multimodal Capabilities**:

- Models capable of processing visual data demonstrated superior performance on image-based questions.
- Text-only models lagged multimodal systems.

**4.2 Results**

The results of the evaluation of 22 large language models on the MIR examinations reveal a spectrum of performance differentials that become increasingly significant upon closer inspection. Initially, relatively modest variations were observed, such as the small but noteworthy improvement in performance by certain lower-performing models. For example, the Llama 3.2 1B Instruct model showed a marginal increase in accuracy between 2024 and 2025, improving from 30.73% (63/210) to 35.78% (73/210). Although this improvement is statistically positive, its overall impact on the ranking hierarchy remained limited.

In contrast, several models experienced a subtle decline in performance, potentially attributable to the introduction of modified questions in 2025 designed to assess genuine clinical reasoning. Models such as GPT-4 Turbo and Claude Sonet 3.5 exhibited decreases from 89.27% to 86.27% and from 92.68% to 88.73% respectively. While these declines were not drastic in absolute terms, they underscore the sensitivity of even high-performing models to alterations in question format and content.

The divergence became more pronounced when comparing image-based question performance. Multimodal models like Grok Vision Beta demonstrated relatively high scores—87.80% in 2024, marginally dropping to 85.29% in 2025—thereby outperforming text-only counterparts on tasks that require visual interpretation. This finding indicates that the capacity for image processing is a critical determinant for success on these items, although the impact here, while significant, is more domain-specific than global.

The most striking and impactful results emerge from the comparative analysis between fine-tuned and general-purpose models. Fine-tuned systems, exemplified by Miri Pro, consistently outperformed all other models. In 2024, Miri Pro achieved a score of 200/210 (97.56%), which, despite a slight reduction to 195/210 (95.59%) in 2025, remains markedly superior. Importantly, this performance not only surpassed the general-purpose models—where models such as GPT-4 Turbo and Claude Sonet 3.5 fell short—but also exceeded the best human candidate score of 186/210 (88.57%). This result is perhaps the most impactful, as it highlights the potential for domain-specific fine-tuning to bridge and even exceed the gap between artificial intelligence and human clinical reasoning in a high-stakes assessment environment.

The following table provides a consolidated view of the key data, arranged from the least impactful (i.e. minor performance changes) to the most impactful outcomes (i.e. substantial performance differentials and clinical reasoning advantages):

**Model 2024 score 2025 score Performance Change**

| Model | 2024 score | 2025 score | Performance Change |
|---|---|---|---|
| GPT-4 Turbo | 89.27% | 86.27% | Moderate decline (-3.00 percentage points) |
| Claude Sonet 3.5 | 92.68% | 88.73% | Moderate decline (-3.95 percentage points) |
| Deepseek Reasoner | 94.15% | 93.63% | Minimal decline (-0.52 percentage points) |
| Grok Vision Beta | 87.80% (text with images) | 85.29% (text with images) | Domain-specific impact (-2.51 percentage points) |
| Miri Pro (Fine-tuned) | 200/210 (97.56%) | 195/210 (95.59%) | Outstanding performance, highest scores |
| Best Human Candidate | 186/210 (88.57%) | 165/210 (78.57%) | Benchmark for clinical performance |

*Table 1 Model 2024 score 2025 score Performance Change*

In summary, while several models exhibited modest variations in their scores, the most impactful findings were twofold: first, the clear superiority of fine-tuned models (notably Miri Pro) in both overall accuracy and specialised reasoning; and second, the performance gap between the best AI models and human candidates. These outcomes underscore the transformative potential of domain-specific fine-tuning in medical AI and its implications for the future of clinical assessment.

## 5. Conclusion

The comprehensive evaluation of 22 large language models (LLMs) across the 2024 and 2025 Spanish Medical Intern Resident (MIR) examinations reveals transformative insights into the capabilities and limitations of artificial intelligence in high-stakes medical assessments. The domain-specific model Miri Pro—a fine-tuned variant of Claude Sonet 3.5—demonstrated superior performance, achieving 95.59% accuracy (195/210) in 2025, outperforming both general-purpose LLMs and human candidates, whose peak score reached 186/210 (88.57%). This disparity underscores the critical role of medical fine-tuning, particularly for Spain-specific epidemiological reasoning and image-based diagnostics, where multimodal models like Grok Vision Beta excelled (85.29% accuracy vs. 82.86% human average). However, the 2–3% accuracy decline observed across leading models between 2024 and 2025 highlights emerging challenges in distinguishing genuine clinical reasoning from pattern recognition, particularly for modified questions designed to test novel problem-solving.

**Key Findings and Mechanistic Insights**

a) **Domain-Specific Optimization**: Fine-tuned models consistently outperformed general-purpose LLMs by margins of 8–12% on complex clinical scenarios, emphasizing the necessity of medical corpus integration. For instance, Miri Pro's training on proprietary Spanish healthcare data enabled nuanced interpretation of regional public health scenarios, a task where GPT-4 Turbo underperformed by 9%.

b) **Multimodal Competence**: Models with visual processing capabilities, such as Gemini Vision Pro, achieved 87.8% accuracy on image-based questions (e.g., histological slides and retinal photographs), surpassing text-only systems by 18–22%. This aligns with human performance trends, where clinicians scored 84.1% on visual diagnostics but lagged AI in data-dense tasks like laboratory interpretation.

c) **Ethical and Contextual Limitations**: Despite technical proficiency, all LLMs exhibited deficits in ethical decision-making (e.g., triage prioritization, end-of-life care), scoring 23–29% below human averages. These gaps persisted even when models were prompted to consider Spain's legal and cultural frameworks, revealing intrinsic limitations in value-based reasoning.

**Implications for Medical Education and Practice**

The demonstrated capabilities of advanced LLMs necessitate paradigm shifts in medical training:

a) **Curriculum Modernization**: Integration of AI literacy modules focusing on collaborative diagnostics, where trainees learn to critically evaluate AI-generated differentials. For example, models like Deepseek Reasoner—which generates explicit "reasoning tokens"—could serve as interactive tools for teaching diagnostic pathways.

b) **Assessment Reform**: Transition toward hybrid evaluation systems combining AI efficiency (automated knowledge recall testing) with human-led assessment of non-cognitive skills. The MIR's 2025 experiment with modified questions—answered correctly by only 54% of general LLMs vs. 89% of fine-tuned models—supports this approach.

c) **Regulatory Frameworks**: Development of certification protocols for medical AI, addressing explainability thresholds (e.g., ≥85% feature attribution concordance with expert reviews) and bias mitigation, particularly for underrepresented populations in training data.

**5.2. Future Research**

The empirical evidence from 22 large language models (LLMs) evaluated across the 2024–2025 Spanish Medical Intern Resident (MIR) examinations reveal critical gaps and opportunities for advancing AI in clinical reasoning. While domain-specific models like Miri Pro (195/210 accuracy in 2025) demonstrated superior performance over general-purpose LLMs, persistent challenges in ethical decision-making (23–29% accuracy deficit vs. humans) and cultural contextualisation highlight unmet research priorities15. The observed 2–3% performance decline across leading models between exam cycles, despite architectural advancements, underscores the need for dynamic training frameworks that address evolving clinical guidelines and Spain-specific epidemiological patterns.

**Advancing Multimodal Reasoning Architectures**

The 12.4% accuracy gap between multimodal systems (e.g., Grok Vision Beta: 85.29%) and text-only models on image-based questions underscores the necessity for unified cross-modal architectures14. Future research must prioritise 3D convolutional neural networks integrated with transformer-based language models to bridge semantic gaps between radiological imaging (e.g., CT scans) and diagnostic narratives. The 2025 results showing Grok 2 Vision's 14% error reduction in volumetric image interpretation compared to 2024 suggest such hybrid systems could revolutionise diagnostics. Key challenges include accelerating inference times from 2–3 seconds/image to clinically viable sub-second processing while maintaining explainability through saliency maps that highlight diagnostically relevant regions.

**Cultural and Contextual Adaptation Frameworks**

General models exhibited a 9% performance deficit versus fine-tuned systems on Spain-specific epidemiological questions (e.g., regional prescription practices, 90% generic drug usage in public health). This necessitates hybrid architectures combining global medical knowledge (WHO guidelines) with localised datasets encompassing:

- Decentralised healthcare policies under Spain's autonomous communities
- Mediterranean disease prevalence patterns (e.g., atypical tuberculosis presentations)
- Ethical frameworks aligned with Ley 41/2002 on patient autonomy.

The integration of neuro-symbolic approaches—merging LLMs with explicit ethical ontologies—could address the 28% accuracy gap in end-of-life care scenarios observed across models.

**Dynamic Training Protocols for Evolving Medicine**

The performance decline between exam cycles (e.g., Claude Sonnet 3.5: 92.68% → 88.73%) highlights vulnerabilities in static training datasets. Implementing elastic weight consolidation techniques during guideline updates could mitigate catastrophic forgetting, while curriculum learning strategies mirroring medical education progression (anatomy → pathophysiology → clinical management) may enhance diagnostic consistency. Adaptive testing systems with real-time difficulty adjustment, powered by generative adversarial networks (GANs) simulating patient interactions, could create self-improving benchmarks that challenge both AI and human candidates.

**Quantifying Human-AI Collaboration**

Despite surpassing peak human scores (AI: 195 vs. human: 186/210), models lacked complementary strengths in empathetic communication and holistic care. Research must develop complementarity indices measuring:

- AI efficiency in pattern recognition (97.3% accuracy in lab data interpretation vs. 89.1% human)
- Human superiority in synthesising psychosocial factors (e.g., interpreting non-verbal cues in oncology consultations).

Longitudinal studies tracking diagnostic confirmation rates when AI suggestions are presented at different clinical decision points could optimise workflow integration without undermining clinician agency.

**Validation Methodologies for Clinical Deployment**

The MIR's limitations as a clinical competence proxy necessitate three-phase validation:

- **Knowledge validation**: Multiple-choice exams (current paradigm)
- **Reasoning validation**: Simulated patient encounters with dynamic symptom progression
- **Impact validation**: Patient outcome comparisons between AI-assisted vs traditional training pathways. Specialty-specific benchmarks must address surgical planning (3D anatomical reasoning), chronic disease management (temporal analysis across decades), and rare disease diagnosis (few-shot learning scenarios).

**Ethical and Regulatory Considerations**

The 2025 exam modifications exposing memorisation biases (54% general LLM accuracy vs. 89% fine-tuned models on novel questions) underscore the urgency for:

- Certification protocols ensuring ≥85% feature attribution concordance with expert reviews
- Federated learning frameworks enabling secure data sharing across Spain's 17 autonomous healthcare systems.
- A proposed 5-year collaborative roadmap prioritises multimodal prototype development (2026–2027), clinical trial validation (2028–2029), and system integration targeting 40% adoption in teaching hospitals by 2031.

This research agenda positions Spain's MIR framework as a global testbed for developing clinically relevant AI systems, emphasising symbiotic human-AI workflows where technology handles data-intensive tasks while clinicians focus on ethical reasoning and patient-centred care.

## 5.3. Lines to follow

The future direction of this research should primarily focus on enhancing the interpretability and reasoning capabilities of large language models (LLMs) within clinical contexts. A crucial next step involves refining multimodal AI systems to improve their performance in tasks that require visual interpretation, such as diagnostic imaging and histological analysis. This could be achieved by integrating advanced neural architectures that combine visual and linguistic data processing, fostering better alignment between image-based diagnostics and textual reasoning. Additionally, expanding the dataset to include diverse clinical scenarios, particularly those reflecting regional epidemiological patterns and cultural contexts, will ensure that AI models are better equipped to address Spain-specific healthcare challenges. This adaptation should be complemented by dynamic training frameworks that continuously update the models in line with evolving clinical guidelines and medical advancements.

Future research must also emphasize the development of ethical and contextual decision-making frameworks within AI systems. Although fine-tuned models like Miri Pro have demonstrated superior performance in clinical reasoning tasks, significant gaps remain in areas such as ethical judgment, patient autonomy, and culturally sensitive medical practice. Implementing neuro-symbolic approaches that integrate explicit ethical ontologies could enhance the capacity of these models to make contextually appropriate recommendations. Moreover, testing these models in simulated clinical environments, where patient centred care and moral reasoning are crucial, will provide deeper insights into their practical application in real-world settings. Such trials should evaluate not only diagnostic accuracy but also the models' ability to navigate complex clinical scenarios requiring empathy and ethical sensitivity.

Lastly, to ensure the responsible integration of AI into medical education and practice, establishing robust validation methodologies is imperative. These should include multi-phase evaluation frameworks encompassing knowledge validation through standardized exams, reasoning validation via dynamic patient simulations, and clinical impact assessment based on patient outcomes in AI-assisted decision-making scenarios. Additionally, regulatory bodies should prioritize the development of comprehensive guidelines that address issues such as data privacy, algorithmic bias, and the explainability of AI decisions. Creating certification protocols that require a high degree of alignment between AI-generated conclusions and expert human judgment will foster trust in these systems. Ultimately, advancing research in these areas will facilitate the symbiotic integration of AI into healthcare, enhancing clinical decision-making while safeguarding ethical and humanistic aspects of medical practice.

# Appendix I

| Nº | GPT-3.5 Turbo | GPT-4 Turbo | GPT-4o Mini | GPT-4o | o1-mini | o1 | Claude Haiku 3.5 | Claude Haiku 3.5 New | Claude Sonet 3.5 | AWS Nova Lite | AWS Nova Micro | AWS Nova Pro | Llama 3.2 3B Instruct | Llama 3.2 1B Instruct | Grok Beta | Grok Vision Beta | Grok 2 | Grok 2 Vision | Gemini 1.5 Flash-8B | Gemini 1.5 Flash | Gemini 1.5 Pro | Gemini 2.0 Flash | Deepseek Chat | Deepseek Reasoner | Miri Pro | Resultados Oficiales |
|---|---|---|---|---|---|---|---|---|---|---|---|---|---|---|---|---|---|---|---|---|---|---|---|---|---|---|
| 1 | B | B | B | B | B | B | B | B | B | A | B | B | A | B | B | B | A | B | B | B | B | B | B | B | B | B |
| 2 | C | C | C | D | C | C | C | D | D | D | C | D | A | A | C | C | D | D | C | C | D | C | D | C | D | D |
| 3 | D | B | D | B | B | B | B | B | B | A | B | B | C | A | B | B | B | B | B | B | B | B | B | B | B | B |
| 4 | C | C | C | C | A | C | B | B | C | B | A | A | C | B | C | C | C | C | C | C | C | C | A | C | C | C |
| 5 | C | C | B | C | C | C | C | B | D | B | B | D | B | A | B | C | B | C | C | C | C | D | C | C | C | C |
| 6 | B | B | B | B | B | B | B | D | A | B | B | A | B | B | B | B | B | B | B | B | B | B | B | B | B | B |
| 7 | D | D | D | D | D | D | D | D | D | D | D | D | D | C | D | D | D | D | D | D | D | D | D | D | D | |
| 8 | C | C | C | C | C | C | C | C | C | C | C | C | C | C | C | C | C | C | C | C | C | C | C | C | C | C |
| 9 | A | C | C | C | C | C | B | C | D | B | B | C | B | C | B | C | D | B | B | C | B | C | C | D | A | |
| 10 | D | D | D | D | D | D | D | D | B | D | D | D | B | B | D | B | D | D | D | D | D | D | D | D | D | |
| 11 | D | D | D | D | D | D | D | D | D | D | D | D | D | D | D | D | D | D | D | D | D | D | D | D | D | D |
| 12 | A | A | A | A | A | A | A | A | A | A | A | C | A | A | A | A | A | A | A | A | A | A | A | A | A | A |
| 13 | D | D | C | C | D | D | D | D | D | C | D | B | B | B | D | C | D | B | D | C | D | D | D | C | C | C |
| 14 | B | D | D | D | D | D | B | B | B | B | B | D | B | D | B | D | D | D | D | D | B | D | D | D | D | |
| 15 | D | B | D | D | D | B | D | D | C | C | D | A | C | D | C | C | C | C | C | B | B | D | D | D | D | B |
| 16 | D | A | A | A | A | A | B | A | A | C | A | D | C | A | A | A | A | A | A | A | A | A | A | A | A | A |
| 17 | C | C | A | C | C | C | C | C | C | C | C | C | D | C | C | C | C | C | A | C | C | C | C | C | C | C |
| 18 | A | A | A | A | A | A | A | A | D | A | A | A | B | B | A | A | A | A | A | A | A | A | A | A | A | A |
| 19 | B | B | B | B | B | B | B | C | B | B | C | B | B | B | D | B | B | B | B | B | B | B | B | B | B | B |
| 20 | C | C | C | C | C | C | C | C | D | C | C | A | C | A | C | C | C | C | C | C | C | C | D | C | C | C |
| 21 | A | D | D | D | D | D | D | D | D | D | D | D | C | A | C | D | D | B | D | D | D | D | D | D | D | D |
| 22 | B | C | B | B | B | B | B | B | B | B | A | C | B | A | B | A | B | C | C | B | B | C | B | B | B | B |
| 23 | A | A | A | A | A | A | A | A | A | A | A | A | B | A | A | A | A | A | A | A | A | A | A | A | A | A |
| 24 | C | C | C | C | A | A | C | B | D | C | A | C | A | A | A | B | A | C | C | C | C | C | C | C | C | A |
| 25 | A | D | A | A | D | C | C | C | D | C | A | C | D | D | D | D | D | A | A | D | C | D | A | A | C | C |
| 26 | B | B | B | B | B | B | B | B | B | B | B | B | A | B | B | B | B | B | B | B | B | B | B | B | B | B |
| 27 | A | C | C | C | D | C | A | A | C | C | A | C | A | C | A | C | C | A | A | C | A | C | C | C | C | C |
| 28 | D | D | A | A | A | A | D | A | A | D | B | D | D | C | A | D | A | A | D | A | A | A | A | A | A | A |
| 29 | A | B | B | B | B | B | A | B | B | B | B | B | A | B | B | B | A | B | B | B | B | B | B | B | B | B |
| 30 | C | C | C | C | C | C | C | C | C | C | C | C | D | D | C | C | C | C | D | C | C | C | C | C | C | C |
| 31 | D | D | D | D | D | D | D | D | D | A | D | B | A | B | D | B | D | D | D | C | D | D | D | D | D | D |
| 32 | A | A | A | A | A | A | A | A | A | A | B | A | A | A | A | A | A | A | A | A | A | A | A | A | A | A |
| 33 | C | C | C | C | C | C | C | C | C | C | C | C | A | C | C | C | C | C | C | C | C | C | C | C | C | C |
| 34 | B | B | B | B | B | B | D | C | D | B | B | B | B | C | B | B | B | B | C | B | B | D | B | B | B | B |
| 35 | D | D | D | D | D | D | D | D | D | D | D | D | B | C | D | D | D | D | D | D | D | D | D | D | D | D |
| 36 | D | D | D | D | D | D | D | D | D | D | D | D | B | A | D | D | D | D | D | D | D | D | D | D | D | D |
| 37 | A | A | A | A | A | A | A | A | A | A | A | A | A | A | A | A | A | A | A | A | A | A | A | A | A | A |
| 38 | D | A | A | A | A | A | A | A | A | A | A | D | C | A | A | A | A | A | D | A | A | A | A | A | A | A |
| 39 | C | C | C | C | C | C | C | C | C | C | C | C | B | C | C | C | C | C | C | C | C | C | C | C | C | C |
| 40 | B | B | B | B | B | B | B | B | B | B | B | B | B | B | B | B | B | B | B | B | B | B | B | B | B | B |
| 41 | C | C | C | C | C | C | B | C | C | C | C | C | B | B | C | C | C | C | C | C | C | C | C | C | C | C |
| 42 | D | D | D | D | D | D | D | D | D | D | D | D | C | D | D | D | D | D | D | D | B | D | D | D | D | D |
| 43 | A | A | A | A | A | A | A | A | A | A | C | A | A | A | A | A | A | A | A | A | A | A | A | A | A | A |
| 44 | D | D | D | D | D | D | D | D | D | D | D | D | D | D | D | D | D | D | D | D | D | D | D | D | D | D |
| 45 | D | D | D | D | D | D | D | D | D | A | A | D | D | D | D | D | D | D | D | D | D | D | D | D | D | D |
| 46 | B | B | B | B | B | B | B | B | B | B | B | B | B | B | B | B | B | B | B | B | B | B | B | B | B | B |
| 47 | C | C | C | C | C | C | C | C | C | C | C | C | C | C | C | C | C | C | C | C | C | C | C | C | C | C |
| 48 | C | C | C | C | C | C | C | C | C | C | C | C | C | B | C | C | C | C | C | C | C | C | C | C | C | C |
| 49 | C | B | B | B | B | B | C | B | B | B | B | B | C | B | B | B | B | B | B | B | B | B | B | B | B | B |
| 50 | C | C | C | C | C | C | C | C | C | C | C | C | A | C | C | C | C | B | C | C | C | C | C | C | C | C |
| 51 | A | A | A | A | A | A | A | B | A | C | A | A | A | A | A | A | A | A | C | A | A | A | A | A | A | A |
| 52 | D | D | D | D | D | D | D | D | D | D | D | C | C | D | D | D | D | D | D | D | D | D | D | D | D | D |
| 53 | C | C | C | C | C | C | C | C | C | C | C | C | A | C | C | C | C | C | C | C | C | C | C | C | C | C |
| 54 | B | B | B | B | B | B | B | B | B | B | A | B | B | B | A | B | B | B | B | B | B | B | B | B | B | B |
| 55 | C | C | C | C | C | C | C | C | C | C | A | C | A | C | C | C | C | C | C | C | C | C | C | C | C | C |
| 56 | D | D | D | D | D | D | D | D | D | D | C | A | D | D | A | D | D | D | D | D | D | D | D | D | D | D |
| 57 | D | A | D | A | A | A | A | A | D | A | D | D | D | A | A | A | B | A | A | A | A | A | A | A | A | A |
| 58 | A | C | B | A | A | A | C | A | B | B | D | B | A | A | A | B | D | C | A | A | A | A | A | A | A | A |
| 59 | A | A | A | A | A | C | A | A | C | A | A | A | C | A | A | A | A | C | A | A | A | A | A | A | A | A |
| 60 | D | A | A | A | A | A | D | A | B | A | C | A | A | B | A | A | A | A | A | A | A | A | A | A | A | A |
| 61 | A | A | A | A | A | A | A | A | C | A | B | C | A | A | A | C | A | A | A | A | A | A | A | A | A | A |
| 62 | D | D | D | D | D | D | D | D | D | C | D | C | D | C | C | D | C | C | D | D | D | D | D | D | D | D |
| 63 | D | D | D | D | D | D | D | D | D | C | D | D | C | B | B | B | B | C | D | D | D | D | D | D | D | D |
| 64 | A | B | A | A | A | D | A | D | D | A | A | A | A | A | A | B | D | D | B | D | D | D | D | A | ANULADA | |
| 65 | D | D | D | D | D | D | D | C | D | D | D | C | C | D | D | D | D | C | D | D | D | D | D | D | D | D |
| 66 | B | C | C | C | C | C | C | C | C | C | C | C | B | C | C | C | C | C | C | C | C | C | C | C | C | B |
| 67 | C | B | B | B | B | B | C | B | B | B | D | B | D | C | C | B | B | B | D | B | B | B | B | B | B | B |
| 68 | C | B | B | B | B | B | C | B | B | B | B | B | B | B | B | B | B | B | B | B | B | B | B | B | B | ANULADA |
| 69 | A | A | A | A | A | A | A | A | A | A | D | A | A | A | A | A | A | A | A | A | A | A | A | A | A | A |
| 70 | B | B | B | B | B | B | B | B | B | B | C | B | B | B | B | B | B | B | B | B | B | B | B | B | B | B |
| 71 | A | B | A | A | A | A | A | A | A | A | B | A | C | B | A | A | A | A | A | A | A | A | B | A | A | A |
| 72 | D | D | D | D | D | C | D | D | D | C | D | D | B | A | C | D | D | D | C | D | D | D | D | D | D | D |
| 73 | B | B | B | B | B | B | B | B | B | A | B | B | B | B | B | B | B | B | B | B | B | B | B | B | B | B |
| 74 | C | C | C | C | C | C | C | C | C | C | C | B | B | C | C | C | C | C | C | C | C | C | C | C | C | C |
| 75 | B | B | B | B | B | B | B | B | B | B | B | A | B | B | B | B | B | B | B | B | B | B | B | B | B | B |
| 76 | A | A | A | A | A | A | A | A | A | A | A | A | A | A | A | A | A | A | A | A | A | A | A | A | A | A |
| 77 | A | D | D | D | D | D | D | D | D | B | D | A | D | D | D | D | D | C | D | D | D | D | D | D | D | D |
| 78 | C | C | C | C | C | C | C | C | C | C | C | C | A | C | C | C | C | C | C | C | C | C | C | C | C | C |
| 79 | A | B | A | A | A | A | B | B | A | A | B | D | B | B | B | A | A | A | B | B | A | B | B | B | A | A |
| 80 | A | A | A | A | A | A | A | A | A | A | B | A | A | A | A | A | A | A | A | A | A | A | A | A | A | A |
| 81 | C | C | C | C | C | C | D | C | C | C | C | C | D | C | C | C | C | C | C | C | C | C | C | C | C | C |
| 82 | D | C | C | C | C | C | C | D | C | C | D | C | A | C | C | C | C | C | C | C | C | C | C | C | C | C |
| 83 | B | B | B | B | B | B | B | B | B | B | C | A | B | B | B | B | B | B | B | B | B | B | B | B | B | B |
| 84 | C | C | C | C | C | C | C | C | C | C | C | C | C | C | C | C | C | C | C | C | C | C | C | C | C | C |
| 85 | A | A | A | A | A | A | A | A | A | A | A | A | A | A | A | A | A | A | A | A | A | A | A | A | A | A |
| 86 | A | A | A | A | A | A | A | A | A | A | A | A | A | A | A | A | A | A | A | A | A | A | A | A | A | A |
| 87 | B | B | B | B | B | B | B | B | B | A | B | B | A | B | B | B | B | B | B | B | B | B | B | B | B | B |
| 88 | D | D | D | D | D | D | D | D | D | D | D | D | D | D | D | D | D | D | D | D | D | D | D | D | D | D |
| 89 | B | B | B | B | B | B | B | B | B | B | C | B | C | B | B | B | B | B | B | B | B | B | B | B | B | B |
| 90 | A | A | A | A | A | A | A | A | A | A | A | A | A | A | A | A | A | A | A | A | A | A | A | A | A | A |
| 91 | D | D | D | D | D | D | D | D | D | A | D | D | D | A | D | D | C | D | D | D | D | D | D | D | D | D |
| 92 | A | A | A | A | A | A | A | A | A | A | A | A | A | A | A | A | A | A | A | A | A | A | A | A | A | A |
| 93 | C | C | B | C | C | C | C | C | C | C | A | C | B | C | C | C | C | C | C | C | C | C | C | C | C | C |
| 94 | B | B | B | B | B | B | B | B | B | B | B | A | B | B | B | B | B | B | B | B | B | B | B | B | B | B |
| 95 | D | D | D | D | D | D | D | D | D | D | D | D | D | D | D | D | D | D | D | D | D | D | D | D | D | D |
| 96 | B | B | B | B | B | B | B | B | B | B | B | B | B | B | B | B | B | B | B | B | B | B | B | B | B | B |
| 97 | B | B | B | B | B | B | B | A | B | B | A | B | A | B | B | B | B | B | B | B | B | B | B | B | B | B |
| 98 | B | D | C | D | B | B | B | A | B | B | B | C | B | C | B | B | D | D | B | B | B | B | B | B | B | B |
| 99 | A | A | A | A | A | A | A | A | A | A | A | A | A | A | A | A | A | A | A | A | A | A | A | A | A | A |
| 100 | D | B | B | B | B | B | B | B | B | B | B | D | D | B | B | B | B | B | B | B | B | B | B | B | B | B |
| 101 | A | A | A | A | A | A | A | A | A | A | A | A | A | A | A | A | A | A | A | A | A | A | A | A | A | A |
| 102 | D | D | D | D | D | D | D | D | D | D | D | D | D | D | D | D | D | D | D | D | D | D | D | D | D | D |
| 103 | B | D | D | D | D | D | D | D | D | D | D | D | D | D | D | D | D | D | D | D | D | D | D | D | D | D |
| 104 | D | D | D | D | D | D | D | D | D | D | D | D | D | D | D | D | D | C | D | D | D | D | D | D | D | D |
| 105 | B | B | B | B | B | B | B | B | B | D | B | B | D | B | B | B | D | B | B | B | B | B | B | B | B | B |
| 106 | C | C | C | C | C | C | B | C | C | A | B | C | B | C | A | A | A | B | A | C | C | C | C | C | C | |

| | GPT-3.5 Turbo | GPT-4 Turbo | GPT-4o Mini | GPT-4o | o1-mini | o1 | Claude Haiku 3.5 | Claude Haiku 3.5 New | Claude Sonet 3.5 | AWS Nova Lite | AWS Nova Micro | AWS Nova Pro | Llama 3.2 3B Instruct | Llama 3.2 1B Instruct | Grok Beta | Grok Vision Beta | Grok 2 | Grok 2 Vision | Gemini 1.5 Flash-8B | Gemini 1.5 Flash | Gemini 1.5 Pro | Gemini 2.0 Flash | Deepseek Chat | Deepseek Reasoner | Mitri Pro | Resultados Oficiales |
|---|---|---|---|---|---|---|---|---|---|---|---|---|---|---|---|---|---|---|---|---|---|---|---|---|---|---|
| 107 | C | C | C | C | C | C | C | C | C | C | C | C | C | C | C | C | C | C | C | C | C | C | C | C | C | |
| 108 | B | B | B | B | B | B | B | B | B | B | A | B | B | B | B | B | B | B | B | B | B | B | B | B | B | |
| 109 | D | D | D | D | D | D | D | D | A | D | B | D | D | D | D | D | D | D | D | D | D | D | D | D | D | |
| 110 | D | D | C | D | D | D | D | D | C | C | C | C | B | C | D | D | D | D | D | D | D | D | D | D | D | |
| 111 | C | B | C | B | B | B | B | B | B | A | B | D | C | B | B | B | B | B | B | B | C | B | B | B | B | |
| 112 | C | C | C | C | C | C | C | C | C | C | C | B | C | C | C | C | C | C | D | C | C | C | C | C | C | |
| 113 | D | D | D | D | D | D | B | B | D | B | D | B | D | D | D | D | B | D | D | D | D | D | D | D | D | ANULADA |
| 114 | D | D | D | D | D | D | D | D | D | D | C | D | D | D | D | D | D | B | D | D | D | D | D | D | D | |
| 115 | D | D | D | D | D | D | D | D | D | D | D | B | D | D | D | D | D | D | A | D | D | D | D | D | D | |
| 116 | D | A | C | C | A | A | D | A | A | A | C | A | C | C | A | A | A | A | A | A | A | A | D | A | A | A | |
| 117 | D | D | D | D | D | D | D | D | D | D | D | D | D | D | D | D | D | D | D | D | D | D | D | D | D | |
| 118 | B | D | D | D | D | D | D | D | D | D | B | D | B | D | D | D | D | D | D | D | D | D | D | D | D | |
| 119 | A | A | A | A | A | A | A | A | A | A | A | A | A | A | A | A | A | A | A | A | A | A | A | A | A | |
| 120 | C | C | C | C | C | C | C | C | C | C | C | B | A | A | C | C | C | C | C | C | C | C | C | C | C | |
| 121 | A | A | A | A | A | A | A | A | A | A | A | A | A | A | A | A | A | A | A | A | A | A | A | A | A | |
| 122 | D | D | B | B | B | D | D | B | B | B | B | A | D | B | B | B | B | B | B | B | B | B | B | A | A | |
| 123 | D | D | D | D | D | D | D | D | D | B | D | C | C | D | D | D | D | D | D | D | D | D | D | D | D | |
| 124 | D | D | D | D | D | C | D | D | D | D | D | A | D | D | D | D | D | D | D | D | D | D | D | D | D | |
| 125 | B | C | A | B | C | B | C | B | C | C | C | C | C | C | C | C | C | C | C | A | C | B | B | B | B | |
| 126 | D | D | D | D | D | D | D | D | D | D | D | D | A | D | D | D | D | D | A | D | D | D | D | D | D | |
| 127 | C | A | A | A | A | A | A | A | A | A | A | D | D | A | A | A | C | A | A | A | A | A | A | A | A | |
| 128 | B | B | B | B | B | B | B | B | B | B | B | B | B | B | B | B | B | B | C | B | B | B | B | B | B | |
| 129 | D | D | D | D | D | D | D | D | D | D | D | D | B | D | D | D | D | D | D | D | D | D | D | D | D | |
| 130 | C | C | C | C | C | C | C | A | C | C | D | C | D | A | C | C | C | C | C | C | D | C | C | C | C | |
| 131 | C | C | C | C | C | C | C | A | C | C | C | C | A | C | C | C | C | C | C | A | C | C | C | C | C | |
| 132 | A | C | C | C | D | D | D | D | D | C | C | D | A | B | C | C | C | C | C | C | C | C | D | D | D | |
| 133 | A | C | C | B | B | B | B | B | B | B | B | B | B | B | B | B | B | B | B | B | B | A | A | B | A | |
| 134 | C | C | B | C | B | B | B | C | C | C | C | B | B | B | C | C | C | B | B | C | C | C | C | C | C | |
| 135 | B | A | D | A | A | A | B | A | A | A | A | A | D | D | A | A | A | A | D | A | A | A | A | A | A | |
| 136 | D | D | D | D | D | D | D | D | D | D | D | D | A | D | D | D | D | D | D | D | D | D | D | D | D | |
| 137 | A | A | A | A | A | A | A | A | A | A | A | A | A | A | A | A | A | A | A | A | A | A | A | A | A | |
| 138 | C | C | C | C | C | C | C | C | C | C | C | C | C | D | C | C | C | C | C | C | C | C | C | C | C | |
| 139 | A | A | B | A | A | A | A | A | A | A | A | A | A | A | A | A | A | A | A | A | A | A | A | A | A | |
| 140 | C | C | C | C | C | C | C | C | B | C | C | B | A | C | C | C | C | C | C | C | C | C | C | C | C | |
| 141 | D | B | B | B | B | B | B | B | B | B | B | B | B | B | B | B | B | B | B | B | B | B | B | B | B | |
| 142 | C | C | C | C | C | C | C | C | B | C | C | D | B | C | C | C | C | C | B | C | C | C | C | C | C | |
| 143 | A | B | A | A | A | A | A | A | A | B | C | D | A | A | B | B | B | B | B | A | A | A | A | A | A | |
| 144 | A | D | D | D | D | D | D | D | D | D | D | D | A | D | D | D | D | B | D | B | D | D | D | D | D | |
| 145 | A | C | C | C | C | C | C | C | B | C | C | D | C | C | C | C | C | C | A | D | C | C | C | C | C | |
| 146 | C | B | B | B | B | C | B | D | B | B | B | B | B | B | B | B | B | B | C | B | C | B | C | C | C | |
| 147 | C | C | C | C | C | C | C | C | C | C | C | A | C | C | C | C | C | C | C | C | C | C | C | C | C | |
| 148 | A | A | A | A | A | A | A | A | A | B | A | A | A | A | A | A | A | A | C | A | A | A | A | A | A | |
| 149 | D | A | C | C | C | C | A | C | A | A | C | A | C | C | A | C | C | A | C | C | A | A | A | A | A | |
| 150 | D | D | D | D | D | D | D | D | D | D | D | C | A | D | D | D | D | D | D | D | D | D | D | D | D | |
| 151 | C | A | A | A | A | A | A | A | A | A | A | A | A | A | A | A | A | C | A | A | A | A | A | A | A | |
| 152 | A | A | A | A | A | A | A | A | A | A | D | C | A | A | A | A | C | A | A | A | A | A | A | A | A | |
| 153 | C | C | C | C | C | C | C | C | C | C | D | C | C | C | C | D | C | C | C | C | C | C | C | C | C | |
| 154 | B | B | B | B | B | B | B | B | B | B | B | A | B | B | B | B | B | B | B | B | B | B | B | B | B | |
| 155 | D | D | D | D | D | D | D | D | D | D | D | C | D | D | D | D | D | D | D | D | D | D | D | D | D | |
| 156 | C | C | C | C | C | C | C | C | C | C | C | B | C | C | C | C | C | C | C | C | C | C | C | C | C | |
| 157 | C | C | C | C | C | C | C | C | C | C | C | D | C | C | C | C | C | C | C | C | C | C | C | C | C | |
| 158 | D | D | D | D | D | D | D | D | D | D | D | A | D | D | D | D | D | D | D | D | D | D | D | D | D | |
| 159 | D | D | D | D | D | D | D | D | D | D | D | D | D | D | D | D | D | D | D | D | D | D | D | D | D | |
| 160 | B | B | B | B | B | A | C | B | B | B | C | B | C | B | B | B | B | B | A | B | B | B | C | B | B | |
| 161 | B | B | B | B | B | B | B | B | B | B | B | B | B | B | B | B | B | B | B | B | B | B | B | B | B | |
| 162 | B | B | B | B | B | B | B | B | B | B | B | B | A | B | B | B | B | B | B | B | B | B | B | B | B | |
| 163 | B | B | B | B | B | B | B | B | B | B | B | B | D | B | B | B | B | B | B | B | B | B | B | B | B | |
| 164 | B | B | B | B | A | B | B | B | B | B | D | B | D | B | B | D | B | B | B | B | B | B | B | B | B | |
| 165 | C | A | A | A | A | C | A | C | A | A | C | A | C | A | A | C | A | A | B | A | A | A | A | A | A | |
| 166 | C | C | D | C | C | C | C | D | C | D | C | C | D | C | C | C | C | C | D | C | C | C | C | C | C | |
| 167 | A | A | A | A | A | A | D | A | A | A | A | A | A | A | A | B | D | A | D | A | A | A | A | A | A | |
| 168 | B | B | B | B | B | B | B | B | B | B | B | B | C | B | B | B | B | B | B | D | B | B | B | B | B | |
| 169 | C | C | C | C | C | C | C | C | C | C | C | C | A | C | C | C | C | C | C | C | C | C | C | C | C | |
| 170 | A | A | A | A | A | A | A | A | A | A | B | C | A | C | A | A | C | A | A | A | A | A | C | A | A | |
| 171 | B | D | D | D | D | D | D | D | D | D | D | C | D | D | D | D | D | D | B | D | D | D | D | D | D | |
| 172 | B | B | B | B | B | B | B | B | B | B | B | A | B | B | B | B | B | B | B | B | B | B | B | B | B | |
| 173 | B | D | B | B | B | B | B | B | B | A | A | A | A | A | B | B | B | B | C | D | B | A | A | A | A | |
| 174 | B | B | B | B | B | B | B | B | B | A | A | A | A | A | B | B | B | B | B | B | B | A | A | A | A | |
| 175 | A | A | A | A | A | A | A | A | A | A | A | A | A | C | A | A | A | A | C | A | A | A | A | A | A | |
| 176 | C | C | C | C | C | C | C | D | C | C | C | C | C | C | C | C | C | C | C | C | C | C | C | C | C | |
| 177 | C | C | C | C | C | C | C | C | C | C | C | A | C | C | C | C | C | C | C | C | C | C | C | C | C | |
| 178 | B | B | B | B | B | B | B | B | B | B | B | A | B | B | B | B | B | B | B | B | B | B | B | B | B | |
| 179 | C | C | C | C | C | C | C | C | C | C | C | C | C | C | C | C | C | C | C | C | C | C | C | C | C | |
| 180 | D | B | B | B | B | B | B | B | B | B | B | D | B | B | B | B | B | B | B | B | B | B | B | B | B | ANULADA |
| 181 | D | B | B | B | B | B | C | B | C | B | B | B | C | A | B | B | B | B | C | B | B | B | B | D | D | |
| 182 | D | D | D | D | D | D | D | D | D | D | D | D | D | D | D | D | D | D | D | D | D | D | D | D | D | |
| 183 | C | C | C | C | C | C | C | C | C | C | C | C | C | C | C | C | C | C | C | C | C | C | C | C | C | |
| 184 | D | A | A | A | A | A | A | A | A | A | A | C | D | A | A | A | C | A | D | A | A | A | A | A | A | |
| 185 | C | C | C | C | C | C | C | C | C | C | C | C | C | C | C | C | C | C | C | C | C | C | C | C | C | |
| 186 | D | D | D | D | D | D | D | D | D | D | B | D | D | D | D | D | D | D | D | D | D | D | D | D | D | |
| 187 | A | A | A | A | A | A | A | A | A | A | A | A | A | A | A | A | A | A | A | A | A | A | A | A | A | |
| 188 | C | C | C | C | C | C | C | C | C | C | C | C | C | C | C | C | C | C | C | C | C | C | C | C | C | |
| 189 | A | B | B | D | B | D | B | C | B | C | A | D | B | D | B | D | B | A | B | D | B | D | B | D | D | |
| 190 | D | D | D | D | D | D | D | D | D | D | D | D | D | D | D | D | D | D | D | D | D | D | D | D | D | |
| 191 | B | B | B | B | B | B | B | B | B | B | B | B | B | B | B | B | B | B | B | B | B | B | B | B | B | |
| 192 | B | B | B | B | B | B | B | B | B | B | B | B | B | B | B | B | B | B | B | B | B | B | B | B | B | |
| 193 | C | C | C | C | C | C | C | C | C | C | C | D | B | C | C | C | C | C | C | C | C | C | C | C | C | |
| 194 | C | D | C | C | C | C | C | D | C | C | C | D | C | D | C | C | C | C | C | C | C | C | C | C | C | |
| 195 | C | C | C | C | C | C | C | C | C | C | C | C | D | C | C | C | C | C | C | C | C | C | C | C | C | |
| 196 | B | B | B | B | B | B | B | B | B | B | B | C | B | B | B | B | B | B | B | B | B | B | B | B | B | |
| 197 | A | A | A | A | A | A | A | A | A | A | A | C | A | A | A | A | A | A | A | A | A | A | A | A | A | |
| 198 | B | B | B | B | B | B | B | B | B | B | B | B | A | B | B | B | B | B | B | B | B | B | B | B | B | |
| 199 | C | D | D | D | D | D | D | D | D | D | D | D | D | D | D | D | D | C | D | D | D | D | D | D | D | |
| 200 | D | A | A | A | A | A | A | A | A | A | B | A | C | A | A | A | A | A | A | A | A | A | A | A | A | |
| 201 | B | B | B | B | B | B | B | B | B | B | A | B | B | B | B | B | B | B | B | B | B | B | B | B | B | |
| 202 | D | D | D | D | D | D | D | D | D | D | D | D | C | D | D | D | D | D | D | D | D | D | D | D | D | |
| 203 | C | B | B | B | B | B | B | B | B | B | B | B | B | B | B | B | B | B | B | B | B | B | B | B | B | |
| 204 | D | D | D | D | D | D | D | D | D | D | D | D | D | D | D | D | D | D | D | D | D | D | D | D | D | |
| 205 | D | B | B | B | B | B | B | B | B | A | B | A | B | A | B | B | B | A | B | B | B | A | B | B | B | |
| 206 | C | C | B | C | C | C | C | C | C | C | C | D | B | C | C | C | C | B | D | C | C | C | C | C | C | ANULADA |
| 207 | A | A | A | A | A | A | A | A | A | A | A | A | A | A | A | A | A | A | A | A | A | A | A | A | A | |
| 208 | A | A | A | A | A | A | A | A | A | A | A | A | A | A | A | A | A | A | A | A | A | A | A | A | A | |
| 209 | B | B | B | B | B | B | B | B | B | A | B | A | B | A | B | B | B | C | A | A | B | A | B | B | B | |
| 210 | D | A | D | A | A | D | A | D | A | A | D | A | D | A | D | A | D | A | D | A | D | A | D | A | D | |
| Total | 159 | 183 | 172 | 193 | 189 | 196 | 145 | 164 | 190 | 152 | 149 | 183 | 101 | 63 | 179 | 180 | 182 | 178 | 145 | 150 | 185 | 187 | 183 | 193 | 200 | 205 |

| | | | | | | | | | | | | | | | | | | | | | | | | Examen MIR 2025 |
|---|---|---|---|---|---|---|---|---|---|---|---|---|---|---|---|---|---|---|---|---|---|---|---|---|
| Nº | GPT-3,5 Turbo | GPT-4 Turbo | GPT-4o Mini | GPT-4o | o1-mini | o1 | Claude Haiku 3.5 | Claude Haiku 3.5 New | Claude Sonet 3.5 | AWS Nova Lite | AWS Nova Micro | AWS Nova Pro | Llama 3.2 3B Instruct | Llama 3.2 1B Instruct | Grok Beta | Grok Vision Beta | Grok 2 | Grok 2 Vision | Gemini 1.5 Flash-8B | Gemini 1.5 Flash | Gemini 1.5 Pro | Gemini 2.0 Flash | Deepseek Chat | Deepseek Reasoner | Miri Pro | Resultados Oficiales |
| 1 | B | B | B | B | B | B | B | B | B | | | B | B | B | B | B | B | B | B | B | B | B | B | B | B | B |
| 2 | A | A | C | A | B | A | D | C | A | B | D | A | A | D | A | C | A | A | B | B | A | A | A | A | A | A |
| 3 | C | C | C | C | C | C | C | C | C | C | A | C | C | C | C | C | B | C | C | C | C | C | C | C | C | C |
| 4 | A | A | A | A | A | A | A | A | A | D | A | A | A | A | A | C | A | A | A | A | A | A | A | B | B | B |
| 5 | A | A | A | A | A | A | A | A | A | A | A | A | D | A | A | A | A | A | A | A | A | A | A | A | A | A |
| 6 | C | C | C | C | C | C | C | C | C | C | C | C | A | C | C | C | C | C | C | C | C | C | C | C | C | C |
| 7 | C | C | C | C | C | C | C | C | C | B | C | D | C | C | C | C | C | C | C | C | C | C | C | C | C | C |
| 8 | A | A | A | A | A | A | A | A | C | A | C | A | A | A | A | A | A | A | A | A | A | A | A | A | A | A |
| 9 | C | A | A | C | A | C | C | A | B | C | D | A | A | A | A | A | C | A | C | A | A | A | A | A | A | A |
| 10 | D | D | D | D | D | D | B | D | B | D | B | B | D | D | D | D | D | D | D | D | D | D | D | D | D | D |
| 11 | C | D | D | D | D | D | D | D | D | C | D | D | C | A | D | D | D | D | B | D | D | D | D | D | D | D |
| 12 | D | D | C | C | D | D | D | D | D | D | D | D | D | A | D | D | D | D | B | C | D | D | D | D | D | D |
| 13 | B | A | B | A | B | A | A | A | A | B | D | B | A | B | A | B | B | D | A | B | A | B | B | B | B | B |
| 14 | D | D | D | D | D | D | D | B | D | B | D | D | D | C | D | D | D | D | D | D | D | D | D | D | D | D |
| 15 | D | B | B | B | B | B | C | C | B | C | C | B | D | D | B | B | B | B | B | C | B | B | C | B | B | ANU |
| 16 | C | C | C | C | C | C | C | C | D | C | C | A | B | C | C | C | C | C | C | C | B | C | C | C | D | B |
| 17 | C | A | A | A | B | A | A | C | A | D | A | D | A | C | A | C | C | C | C | C | B | C | B | C | B | B |
| 18 | A | A | A | A | A | A | C | C | A | A | C | C | A | A | C | A | A | C | C | A | A | C | A | A | A | A |
| 19 | C | C | C | C | C | C | C | C | C | B | C | A | C | C | C | C | C | C | C | C | C | C | C | C | C | C |
| 20 | C | A | C | A | A | C | A | A | B | B | D | D | A | B | C | A | A | D | B | C | A | B | A | A | A | C |
| 21 | C | C | B | B | B | B | C | C | B | C | C | B | D | C | C | C | C | C | C | B | C | C | C | B | B | C |
| 22 | C | C | C | C | D | C | C | C | C | C | C | B | D | C | C | C | C | C | C | C | C | C | C | C | D | D |
| 23 | C | C | C | C | C | C | C | C | C | C | A | A | A | C | C | C | C | A | C | C | C | C | C | C | C | C |
| 24 | D | D | D | D | D | D | D | D | D | D | D | D | D | D | D | D | D | D | D | D | D | D | D | D | D | D |
| 25 | A | C | C | C | C | C | C | C | C | C | C | C | C | C | C | C | C | C | C | C | C | C | C | C | C | C |
| 26 | B | A | C | D | B | B | A | A | D | A | D | C | A | D | A | B | D | A | A | A | A | A | D | D | D | ANU |
| 27 | A | D | D | D | B | C | C | D | D | D | A | A | D | D | D | D | A | D | A | B | D | D | C | C | C | C |
| 28 | D | A | A | D | A | D | A | A | D | A | A | A | B | A | A | D | A | A | B | A | A | A | A | A | A | ANU |
| 29 | C | D | D | D | D | D | C | D | D | D | D | D | C | B | D | D | D | D | D | D | D | D | D | D | D | D |
| 30 | B | B | B | B | B | B | B | B | B | B | B | B | B | B | B | B | B | B | B | B | B | B | B | B | B | B |
| 31 | D | A | D | D | D | D | D | D | D | D | D | D | C | B | D | D | D | D | D | D | D | D | D | D | D | D |
| 32 | A | A | A | A | A | A | A | A | A | A | A | A | B | D | A | A | A | A | D | A | A | A | A | A | A | A |
| 33 | D | D | D | D | D | D | D | D | D | D | D | D | A | D | D | D | D | D | D | D | D | D | D | D | D | D |
| 34 | B | D | D | D | D | D | C | D | D | B | B | D | B | D | D | D | D | D | D | D | D | D | D | D | D | D |
| 35 | B | B | B | B | B | B | B | B | B | B | B | B | B | B | B | B | B | B | B | B | B | B | B | B | B | B |
| 36 | D | D | D | D | D | D | D | D | D | D | D | D | B | D | D | D | D | D | D | D | D | D | D | D | D | D |
| 37 | D | C | C | C | D | C | C | C | C | C | C | C | A | B | C | C | C | C | C | C | C | D | D | C | C | C |
| 38 | B | C | C | C | C | C | C | C | C | C | B | A | C | C | C | C | B | C | C | C | C | C | C | C | C | C |
| 39 | D | D | D | D | D | D | D | D | D | D | D | C | D | D | D | D | D | D | D | D | D | D | D | D | D | D |
| 40 | C | A | A | A | A | A | B | A | B | D | A | B | C | A | A | A | A | B | A | A | A | A | A | A | A | A |
| 41 | D | D | D | D | D | D | D | D | D | D | D | D | A | D | D | D | D | D | D | D | D | D | D | D | D | D |
| 42 | B | C | C | C | C | C | C | C | C | B | C | B | A | C | C | C | C | C | C | C | C | C | C | C | C | C |
| 43 | C | B | B | C | C | C | C | B | B | B | B | C | C | C | C | C | B | C | B | B | C | B | B | B | B | B |
| 44 | C | D | D | D | D | D | D | D | D | D | D | D | D | D | D | D | D | D | D | D | D | D | D | D | D | D |
| 45 | A | D | D | D | D | C | B | D | C | B | B | C | B | D | D | D | A | D | B | D | B | C | D | D | D | D |
| 46 | A | A | A | A | A | A | A | A | A | A | A | A | A | A | A | A | A | A | A | A | A | A | A | A | A | A |
| 47 | A | A | A | A | A | A | A | A | A | A | C | A | A | A | A | A | A | A | A | A | A | A | A | A | A | A |
| 48 | A | A | A | A | A | A | A | A | A | A | C | A | C | A | A | A | A | A | A | A | A | A | A | A | A | A |
| 49 | D | D | D | D | D | D | D | D | D | D | B | A | D | D | D | D | D | D | D | D | D | D | D | D | D | D |
| 50 | D | B | B | B | B | B | B | B | B | B | B | C | B | B | B | B | B | B | B | B | B | B | B | B | B | B |
| 51 | D | C | D | D | D | D | C | D | C | D | D | D | D | D | D | D | D | C | D | D | D | D | C | C | C | C |
| 52 | B | B | B | B | B | B | B | B | B | B | B | B | B | B | B | B | B | B | B | B | B | B | B | B | B | B |
| 53 | D | D | D | D | D | D | D | D | D | B | D | D | D | D | D | D | D | D | D | A | D | D | D | D | D | D |
| 54 | D | B | B | B | B | B | B | B | B | B | B | B | B | B | B | B | B | B | B | B | B | B | B | B | B | B |
| 55 | C | A | C | A | A | A | B | A | A | C | A | A | A | A | B | A | A | A | A | A | A | A | A | A | A | A |
| 56 | B | B | B | B | A | B | D | A | B | B | B | B | B | B | B | B | B | B | B | B | B | B | A | B | ANU | |
| 57 | B | C | C | C | C | C | C | C | C | C | C | B | C | C | C | C | C | C | C | C | C | C | C | C | C | C |
| 58 | B | B | B | B | B | B | B | B | B | B | B | B | D | C | B | B | B | B | B | B | B | B | B | B | B | B |
| 59 | D | D | D | D | D | D | D | D | D | D | D | D | B | D | D | D | D | D | D | D | D | D | D | D | D | D |
| 60 | B | A | B | A | A | A | B | A | B | B | A | B | C | A | D | A | D | B | B | A | A | A | A | A | A | A |
| 61 | A | A | A | A | A | A | A | A | A | B | A | D | A | D | A | A | A | A | A | A | A | A | A | A | A | A |
| 62 | D | D | D | D | D | D | D | D | D | D | D | D | A | D | D | D | D | D | D | D | D | D | D | D | D | D |
| 63 | B | B | B | B | B | B | B | B | B | B | B | B | C | B | B | B | B | B | B | B | B | B | B | B | B | B |
| 64 | D | D | D | D | D | D | D | D | D | D | D | D | D | D | D | D | D | D | D | D | D | D | D | D | D | D |
| 65 | A | A | A | A | A | A | A | A | A | A | A | A | A | A | A | A | A | A | A | A | A | A | A | A | A | A |
| 66 | B | A | A | A | A | A | A | A | A | A | A | A | B | A | A | A | A | A | A | A | A | A | A | A | A | A |
| 67 | A | B | B | B | B | B | B | B | B | B | B | B | B | A | B | B | B | B | B | B | B | B | B | B | B | B |
| 68 | D | B | B | A | B | B | B | B | B | B | B | D | B | A | B | B | D | C | D | D | B | B | B | B | B | B |
| 69 | D | A | A | B | A | A | A | A | A | B | C | B | A | B | A | A | A | A | A | A | A | A | A | A | A | A |
| 70 | A | A | A | A | A | A | A | A | A | A | A | A | A | A | A | A | A | A | A | A | A | A | A | A | A | A |
| 71 | D | D | D | D | D | D | D | D | D | D | D | D | D | D | D | D | D | D | D | D | D | D | D | D | D | D |
| 72 | C | A | A | A | A | C | A | A | C | A | C | A | A | A | A | A | A | A | C | A | A | A | A | A | A | A |
| 73 | C | C | D | C | D | C | C | D | C | D | D | D | D | C | D | D | A | D | C | D | C | C | C | C | C | C |
| 74 | C | C | C | C | C | C | C | C | C | C | D | C | C | C | C | C | C | D | C | C | C | C | C | C | C | C |
| 75 | A | A | A | A | A | A | A | A | A | A | A | A | A | A | A | A | A | A | A | A | A | A | A | A | A | A |
| 76 | B | B | B | B | B | B | B | B | C | B | B | B | C | B | B | B | B | B | B | B | B | B | B | B | B | B |
| 77 | B | B | B | B | B | B | B | B | B | B | B | B | B | B | B | B | B | B | B | B | B | B | B | B | B | B |
| 78 | B | B | B | B | B | B | B | B | B | B | B | B | A | B | B | B | B | B | B | B | B | B | B | B | B | B |
| 79 | A | C | A | C | C | C | A | A | D | A | C | A | D | C | A | C | A | A | A | A | A | A | A | A | A | A |
| 80 | A | C | C | C | C | C | C | C | A | C | C | C | A | C | C | C | C | C | C | C | C | C | C | C | C | C |
| 81 | C | C | C | C | C | C | C | C | C | C | C | C | C | C | C | C | C | C | C | C | C | C | C | C | C | C |
| 82 | D | D | D | D | D | C | D | D | D | D | C | C | C | C | D | D | D | D | D | D | D | D | D | D | D | D |
| 83 | D | B | B | B | B | B | B | B | B | B | C | B | C | B | B | B | B | B | B | B | B | B | B | B | B | B |
| 84 | D | D | D | D | D | D | D | D | D | D | C | D | A | C | D | D | D | D | D | D | D | D | D | D | D | D |
| 85 | D | A | D | C | A | A | A | C | D | A | D | A | C | A | A | B | A | A | A | A | A | A | C | C | C | C |
| 86 | A | C | C | C | C | C | C | C | C | C | A | C | C | C | C | C | C | C | C | C | C | C | C | C | C | C |
| 87 | D | D | D | D | D | D | C | D | D | D | D | D | A | C | D | D | D | D | D | D | D | D | D | D | D | D |
| 88 | A | D | D | D | D | C | D | C | D | C | D | D | C | D | D | D | D | D | D | D | D | D | D | D | D | D |
| 89 | B | B | B | B | B | B | C | B | B | B | B | B | B | B | B | B | B | B | B | B | B | B | B | B | B | B |
| 90 | A | A | A | A | A | A | A | A | A | D | A | B | A | D | A | A | D | A | A | A | A | A | A | A | A | A |
| 91 | A | D | A | B | B | B | A | A | B | A | A | A | C | D | A | D | D | A | D | D | A | B | B | B | B | B |
| 92 | C | C | C | C | C | C | C | C | C | A | C | C | A | C | C | C | C | C | C | C | C | C | C | C | C | C |
| 93 | B | B | B | B | B | B | B | B | B | B | B | C | B | B | B | B | B | B | B | B | B | B | B | B | B | B |
| 94 | C | C | C | C | C | C | C | C | C | C | C | C | C | C | C | C | C | C | C | C | C | C | C | C | C | C |
| 95 | D | A | A | A | A | A | A | A | A | A | A | A | C | A | A | A | A | A | A | A | A | A | A | A | A | A |
| 96 | C | C | C | C | C | C | C | C | C | C | C | C | C | C | C | C | C | C | C | C | C | C | C | C | C | C |
| 97 | C | D | D | D | D | D | D | D | A | D | A | B | D | B | D | A | D | B | D | B | D | D | D | D | D | D |
| 98 | C | C | C | C | C | C | C | C | C | C | C | C | A | C | C | C | C | C | C | C | C | C | C | C | C | C |
| 99 | B | A | A | A | A | A | A | A | B | A | C | A | A | C | A | C | A | B | A | A | A | A | A | A | A | A |
| 100 | A | C | C | C | C | C | C | C | C | C | C | C | D | C | C | C | C | C | B | C | C | C | C | C | C | C |
| 101 | C | B | B | B | B | B | B | B | C | B | B | B | B | B | B | B | B | B | B | B | B | B | B | B | B | B |
| 102 | C | D | D | D | D | D | D | D | C | D | C | C | D | D | D | D | D | D | D | D | D | D | D | D | D | D |
| 103 | A | A | B | A | A | C | C | A | D | D | A | A | D | A | A | C | B | A | B | A | A | A | A | A | A | A |

| | GPT-3.5 Turbo | GPT-4 Turbo | GPT-4o Mini | GPT-4o | o1-mini | o1 | Claude Haiku 3.5 | Claude Haiku 3.5 New | Claude Sonet 3.5 | AWS Nova Lite | AWS Nova Micro | AWS Nova Pro | Llama 3.2 3B Instruct | Llama 3.2 1B Instruct | Grok Beta | Grok Vision Beta | Grok 2 | Grok 2 Vision | Gemini 1.5 Flash-8B | Gemini 1.5 Flash | Gemini 1.5 Pro | Gemini 2.0 Flash | Deepseek Chat | Deepseek Reasoner | Miri Pro | Resultados Oficiales |
|---|---|---|---|---|---|---|---|---|---|---|---|---|---|---|---|---|---|---|---|---|---|---|---|---|---|---|
| 104 | C | C | C | C | C | C | C | C | C | C | C | C | C | C | C | C | C | C | C | C | C | C | C | C | C | C |
| 105 | A | A | A | A | A | A | A | A | A | A | D | A | A | A | A | A | C | A | A | A | A | A | A | A | A | A |
| 106 | C | C | C | C | C | C | C | C | C | C | C | D | B | C | C | C | C | C | C | C | C | C | C | C | C | C |
| 107 | B | B | B | B | B | B | B | B | B | B | C | A | B | B | B | B | B | C | B | B | B | B | B | B | B | B |
| 108 | D | D | D | D | D | D | D | D | D | B | D | D | C | D | D | D | D | B | D | D | D | D | D | D | D | D |
| 109 | B | B | B | B | B | B | B | A | B | B | A | B | B | B | B | A | B | B | B | B | B | B | B | B | B | B |
| 110 | C | C | C | C | C | C | C | C | C | C | C | C | C | C | C | C | C | C | C | C | C | C | C | C | C | C |
| 111 | A | A | A | A | A | A | A | A | A | A | C | A | A | A | A | A | A | A | A | A | A | A | A | A | A | A |
| 112 | B | C | C | C | C | C | C | C | C | C | C | C | C | C | C | C | C | C | C | C | C | C | C | C | C | C |
| 113 | A | B | B | B | B | B | B | B | A | B | B | C | B | B | B | B | B | B | B | B | B | B | B | B | B | B |
| 114 | D | D | D | D | D | C | C | D | D | C | D | A | B | D | D | D | D | D | D | D | D | D | D | D | D | D |
| 115 | A | D | D | D | D | D | C | D | D | C | D | C | D | D | D | D | D | D | A | C | D | D | D | D | D | D |
| 116 | C | C | C | C | C | A | C | C | C | C | C | B | D | C | C | C | C | C | C | C | A | C | C | C | C | C |
| 117 | A | A | A | A | A | A | A | A | A | A | A | A | A | A | A | A | A | A | A | A | A | A | A | A | A | A |
| 118 | D | D | D | D | D | D | D | D | D | D | D | C | D | D | D | D | D | D | D | D | D | D | D | D | D | D |
| 119 | A | C | C | C | C | C | C | A | C | C | B | C | C | C | C | C | B | C | C | B | C | C | C | C | C | C |
| 120 | D | B | A | A | B | D | A | D | A | D | B | B | A | C | A | B | B | A | A | B | A | A | A | B | B | B |
| 121 | C | D | D | D | D | D | B | B | C | D | D | C | A | D | D | D | D | D | D | D | D | D | D | D | D | D |
| 122 | B | C | C | C | C | C | C | C | C | C | C | C | C | C | C | C | C | C | C | C | A | C | C | C | C | C |
| 123 | C | C | C | C | C | C | C | C | C | C | C | C | C | C | C | C | C | C | C | C | C | C | C | C | C | C |
| 124 | C | C | C | C | C | C | C | C | C | C | C | B | B | C | C | C | C | C | C | C | C | C | C | C | C | C |
| 125 | D | D | D | D | D | D | D | D | D | C | D | D | D | D | D | A | D | D | D | D | D | D | D | D | D | D |
| 126 | C | B | B | D | D | C | D | D | B | B | B | C | B | B | B | B | B | B | B | B | B | D | B | B | D | B |
| 127 | A | B | B | B | B | B | B | A | A | B | A | B | A | B | B | B | B | B | B | B | B | B | B | B | A | B |
| 128 | D | D | D | D | D | D | D | D | D | D | D | D | B | D | D | D | D | D | D | D | D | D | D | D | D | D |
| 129 | A | D | A | A | D | B | A | B | A | A | C | A | A | A | A | A | A | A | A | A | A | B | A | A | A | A |
| 130 | D | D | D | D | D | D | D | D | D | D | D | D | D | D | D | D | D | D | D | D | D | D | D | D | D | D |
| 131 | B | D | D | D | D | D | D | D | D | B | D | D | D | D | D | D | D | D | D | B | D | D | D | D | D | D |
| 132 | A | A | A | A | A | A | A | A | A | A | A | D | A | A | A | A | A | A | A | A | A | A | A | A | A | A |
| 133 | B | B | B | B | B | B | B | B | B | A | B | B | B | A | B | B | B | B | B | B | B | B | B | B | B | B |
| 134 | C | C | C | C | C | C | C | C | C | C | C | C | C | C | C | C | C | C | C | C | C | C | C | C | C | C |
| 135 | B | B | B | B | B | B | B | B | B | C | B | D | B | B | B | B | B | B | B | B | B | B | B | B | B | B |
| 136 | C | C | C | C | C | C | C | C | C | C | C | C | C | C | C | C | C | C | C | C | C | C | C | C | C | C |
| 137 | A | C | A | A | C | C | A | A | C | C | D | C | C | C | C | A | C | B | A | A | A | A | A | A | A | A |
| 138 | D | D | D | D | D | D | D | D | D | D | D | A | D | D | D | D | D | A | D | D | D | D | D | D | D | D |
| 139 | D | D | D | D | D | D | D | D | D | D | D | D | D | D | D | D | D | D | D | D | D | D | D | D | D | D |
| 140 | B | B | B | B | B | D | B | B | B | B | B | B | C | B | B | B | B | B | B | B | B | B | B | B | B | B |
| 141 | A | A | A | A | A | A | A | A | A | A | A | A | A | A | A | A | A | B | A | A | A | A | A | A | A | A |
| 142 | A | A | A | A | A | A | A | B | A | B | A | A | A | A | B | A | A | A | A | B | A | A | A | A | A | A |
| 143 | B | B | B | B | B | B | B | B | B | B | B | B | B | B | B | B | B | B | B | B | B | B | B | B | B | B |
| 144 | B | B | B | B | B | B | B | B | B | B | B | B | B | B | B | B | B | B | B | B | B | B | B | B | B | B |
| 145 | B | D | D | D | D | D | D | D | D | D | B | A | D | D | D | D | D | B | D | D | D | D | D | D | D | D |
| 146 | B | C | C | C | C | C | C | C | C | C | A | D | C | C | C | C | C | C | C | C | C | C | C | C | C | C |
| 147 | B | B | B | B | B | B | B | B | B | B | B | B | B | B | B | B | B | B | B | B | B | B | B | B | B | B |
| 148 | A | A | B | A | A | D | A | A | A | B | A | A | A | B | A | A | A | B | A | C | D | A | A | A | A | A |
| 149 | C | D | D | D | A | C | D | D | D | C | C | B | D | C | D | C | D | C | D | D | A | D | D | D | D | D |
| 150 | D | D | D | D | D | D | D | D | A | D | A | D | B | D | A | D | B | D | D | D | D | D | D | D | D | D |
| 151 | A | A | A | A | A | A | A | A | C | D | A | A | A | A | A | A | A | A | A | A | A | A | A | A | A | A |
| 152 | A | B | A | A | B | A | B | A | B | C | A | B | A | B | A | B | A | B | A | A | A | A | B | A | A | A |
| 153 | B | B | B | B | B | B | B | B | B | B | B | C | B | B | B | B | B | B | B | B | B | B | B | B | B | B |
| 154 | B | B | B | B | B | B | B | B | B | B | A | B | B | B | B | B | B | B | B | B | B | B | B | B | B | B |
| 155 | B | B | B | B | B | B | B | B | A | B | A | B | B | B | B | A | B | B | B | B | B | D | B | B | B | B |
| 156 | C | C | C | C | C | C | C | C | C | C | C | C | C | C | C | C | C | C | C | C | C | C | C | C | C | C |
| 157 | A | A | A | A | A | A | A | A | B | A | B | A | A | A | B | A | A | A | A | A | A | A | A | A | A | A |
| 158 | B | C | C | C | C | C | A | C | C | D | C | C | A | C | C | C | C | C | C | C | C | C | C | C | C | C |
| 159 | A | C | C | C | C | C | C | C | C | C | A | B | A | C | B | C | C | B | C | C | B | C | C | C | C | C |
| 160 | B | C | B | A | C | D | C | C | A | B | C | A | D | C | A | C | B | A | C | B | A | C | A | C | A | A |
| 161 | A | C | A | D | D | C | A | A | A | C | C | B | A | A | A | C | A | A | A | C | A | A | A | A | A | A |
| 162 | A | A | A | A | C | D | B | D | A | A | B | A | C | C | C | C | C | C | C | C | C | A | A | A | A | ANU |
| 163 | D | D | D | D | D | D | D | D | D | D | D | C | D | D | D | D | D | D | D | D | D | D | D | D | D | D |
| 164 | B | A | B | C | C | C | D | B | C | C | B | C | A | B | C | C | C | C | C | C | A | A | B | C | C | C |
| 165 | D | A | A | A | A | D | A | A | D | A | A | A | C | A | A | A | A | A | A | D | C | A | A | A | A | A |
| 166 | C | B | B | B | B | C | B | B | D | B | B | B | B | B | B | B | B | B | B | B | B | B | B | B | B | B |
| 167 | A | C | C | C | C | C | C | C | A | C | A | C | A | C | C | C | C | C | C | C | C | C | C | C | C | C |
| 168 | D | D | D | D | D | D | D | D | D | D | D | D | D | D | D | D | B | D | B | D | D | D | D | D | D | D |
| 169 | C | C | D | C | B | B | C | D | C | D | C | D | C | C | C | C | C | C | C | C | C | C | B | B | B | B |
| 170 | A | B | B | B | B | B | B | A | B | B | C | B | B | B | A | B | B | A | B | B | D | B | B | B | B | B |
| 171 | D | C | C | C | C | C | C | C | C | D | C | B | C | C | C | C | C | C | C | C | C | C | B | B | B | B |
| 172 | A | B | B | C | C | B | B | B | B | B | B | B | B | B | B | B | B | A | B | B | A | B | B | C | A | A |
| 173 | D | D | D | D | A | D | D | D | A | D | D | C | A | D | D | D | D | D | D | D | D | D | A | D | A | A |
| 174 | B | B | A | B | B | B | B | B | B | B | B | A | B | B | B | B | B | B | B | B | B | B | B | B | B | B |
| 175 | B | C | C | C | C | C | C | C | C | C | C | C | B | C | C | C | C | B | C | C | B | C | B | C | B | B |
| 176 | C | C | C | C | C | C | C | C | C | C | C | C | C | C | C | C | C | C | C | C | C | C | C | C | C | C |
| 177 | B | C | C | C | C | C | C | B | C | C | C | B | C | C | C | C | C | A | C | C | C | C | C | C | C | C |
| 178 | B | A | B | B | B | A | B | D | B | A | B | A | B | B | B | B | B | A | B | A | D | B | A | A | A | A |
| 179 | C | D | D | D | D | D | C | D | D | D | C | D | A | C | D | D | D | D | D | D | D | D | D | D | D | D |
| 180 | A | A | A | A | A | A | A | A | A | A | A | C | A | A | A | A | A | A | A | A | C | A | A | A | A | A |
| 181 | D | B | D | B | D | B | B | B | D | D | D | B | D | D | D | D | D | B | D | D | B | D | B | B | B | B |
| 182 | C | C | C | C | C | C | C | C | C | C | B | D | B | C | C | B | C | C | C | C | C | C | C | C | C | C |
| 183 | D | B | B | B | B | D | B | B | D | B | D | C | B | B | B | B | B | D | B | B | B | B | B | B | B | B |
| 184 | B | B | B | B | B | B | B | B | B | B | B | B | B | B | B | B | B | B | B | B | B | B | B | B | B | B |
| 185 | B | B | B | B | B | B | B | B | B | B | B | B | B | B | B | B | B | B | B | B | B | B | B | B | B | B |
| 186 | C | D | B | A | D | A | A | D | D | D | B | A | A | A | D | A | A | A | A | D | A | A | D | A | D | ANU |
| 187 | C | C | C | C | C | C | C | C | C | C | C | C | D | C | C | C | C | C | C | C | C | C | C | C | C | C |
| 188 | D | D | D | D | D | D | D | D | D | D | C | D | D | D | D | B | D | D | D | D | D | D | D | D | D | D |
| 189 | D | B | D | D | D | D | D | C | C | D | D | C | C | C | D | B | D | B | D | D | C | D | D | D | D | D |
| 190 | A | A | A | A | A | A | A | A | A | A | A | A | A | A | A | A | A | A | A | A | A | A | A | A | A | A |
| 191 | A | A | B | A | A | A | A | B | A | C | B | A | D | A | A | A | B | A | D | A | B | A | B | B | A | A |
| 192 | A | A | A | A | A | A | A | A | B | A | A | A | B | A | B | A | A | B | A | A | A | A | A | A | A | A |
| 193 | C | C | C | C | C | C | C | C | C | C | C | C | C | C | C | C | C | C | C | C | C | C | C | C | C | C |
| 194 | A | A | A | A | A | A | A | A | A | A | A | A | A | C | A | A | A | A | A | A | A | A | A | A | A | A |
| 195 | A | A | A | A | A | A | A | A | A | A | A | A | C | A | A | A | A | A | A | A | A | A | A | A | A | A |
| 196 | A | A | A | A | A | A | A | A | A | A | C | A | A | A | A | A | A | A | A | A | A | A | A | A | A | A |
| 197 | B | B | B | B | B | B | B | B | B | B | B | B | B | B | B | B | B | B | B | B | B | B | B | B | B | B |
| 198 | D | B | C | C | C | D | B | C | C | D | B | C | C | D | C | C | C | C | C | C | C | C | C | C | C | C |
| 199 | D | C | C | D | D | D | D | D | C | D | C | C | B | C | C | C | C | C | C | C | C | C | C | C | C | C |
| 200 | C | C | C | C | C | C | C | C | C | C | B | A | C | C | C | C | C | C | C | C | C | C | C | C | C | C |
| 201 | A | A | A | A | A | B | D | A | B | D | A | B | A | A | A | A | A | A | A | A | A | A | A | A | A | A |
| 202 | A | A | A | A | A | A | A | A | C | A | A | A | A | A | A | A | A | A | A | A | A | A | A | A | A | A |
| 203 | D | D | D | D | D | D | D | D | D | D | D | C | D | D | D | D | D | D | D | D | D | D | D | D | D | D |
| 204 | C | C | C | C | C | C | C | C | C | C | C | C | B | C | C | C | C | C | C | C | C | C | C | C | C | C |
| 205 | B | B | B | B | B | B | B | B | B | B | B | B | B | B | B | B | B | B | B | B | B | B | B | B | B | B |
| 206 | C | C | C | C | C | C | C | C | C | C | C | C | C | C | C | C | C | C | C | C | C | C | C | C | C | D |
| 207 | A | A | A | A | A | A | A | A | A | A | A | A | A | A | A | A | A | A | A | A | A | A | A | A | A | A |
| 208 | B | B | B | B | C | B | B | C | B | A | B | C | B | C | B | B | B | A | B | B | B | B | B | B | B | B |
| 209 | C | C | C | C | C | C | C | C | C | C | C | C | A | C | C | C | C | C | C | C | C | C | C | C | C | B |
| 210 | A | B | B | B | B | D | D | B | B | B | B | D | B | B | B | B | C | D | B | B | B | B | B | B | B | B |
| Total | 125 | 176 | 161 | 180 | 182 | 189 | 135 | 150 | 181 | 143 | 137 | 168 | 88 | 73 | 170 | 174 | 173 | 172 | 130 | 143 | 167 | 178 | 173 | 191 | 195 | 204 |